\documentclass[11pt]{article}

\usepackage[final]{acl}
\usepackage{times}
\usepackage{latexsym}
\usepackage{amsmath,amssymb}
\usepackage[T1]{fontenc}
\usepackage[utf8]{inputenc}
\usepackage{microtype}
\usepackage{graphicx}
\usepackage{booktabs}
\usepackage{tabularx}
\usepackage{array}
\usepackage{xspace}

\usepackage{xurl}

\newcommand{\benchmark}{CoLT-Drive\xspace}
\newcommand{\method}{KPA\xspace}
\newcommand{\regmoe}{\mbox{RegMoE}\xspace}

\newcolumntype{Y}{>{\raggedright\arraybackslash}X}

\title{CoLT-Drive: Counterfactual Long-Tail Benchmarking and Knowledge-Preserving Adaptation for Driving Affordance Prediction}

\author{Zhengxu Tang, Guofeng Cui, Ziyu Gong, Xiaozhou Zhang, \\
  \bfseries Ruifeng Deng, Chengzhi Qi, Ke Chen, Sachin Patil, \\
  \bfseries Tianjun Xiao, Langechuan Liu, Pichao Wang \\
  NVIDIA \\}

\begin{document}
\maketitle

\begin{abstract}
Long-tail autonomous driving failures are often framed as rare-object recognition errors. We argue that this view is incomplete: the decision-critical question is not only whether a model recognizes an unusual object, but whether it infers how that object changes the ego vehicle's feasible high-level actions. We formalize this problem as \emph{decision-level driving affordance prediction}, where a model maps a front-view image, ego-motion history, and navigation command to a structured longitudinal--lateral meta-action. To evaluate this capability, we introduce \benchmark, a 3,536-sample counterfactual long-tail benchmark that inserts rare objects into otherwise fixed driving scenes and measures whether models predict acceptable action pairs. To improve deployable small VLMs, we propose \method, a knowledge-preserving adaptation framework that combines structured perception-to-decision prompting, SLERP-based expert merging, and \regmoe, a regime-aware LoRA mixture-of-experts module. \method preserves the pretrained model's open-world knowledge while allocating lightweight adaptation capacity to different driving decision regimes. Experiments on an in-domain driving split and \benchmark show that \method achieves 60.8\% pair accuracy on \benchmark, outperforming the pretrained Qwen3-VL-2B baseline (50.3\%) and LoRA SFT (32.4\%) while maintaining competitive in-domain accuracy. Our benchmark and code are available at \url{https://huggingface.co/datasets/tangzx2024/CoLT-Drive} and \url{https://github.com/tangzhengxu/CoLT-Drive}.

\end{abstract}

\section{Introduction}
\label{sec:intro}

Autonomous driving systems have become increasingly reliable in frequent, well-instrumented traffic scenarios, yet they remain brittle under long-tail corner cases involving rare or unusual objects~\citep{coda}. Such failures are often studied through object-level corner-case detection or self-driving VLM understanding~\citep{coda,codalm}. We argue that this view is incomplete: in driving, recognizing an object is only an intermediate step. The decision-relevant question is whether the object changes what the ego vehicle can or should do.

\begin{figure}[t]
    \centering
    \includegraphics[width=0.98\linewidth]{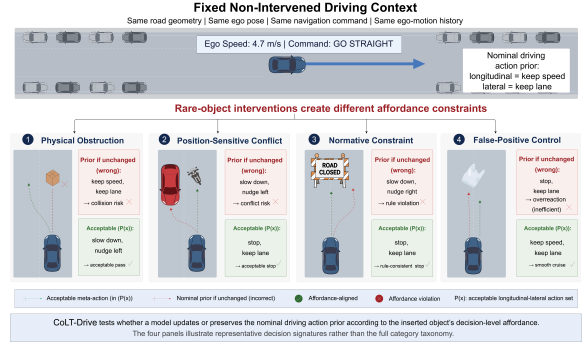}
    \caption{
    Conceptual illustration of \benchmark: fixed driving context, controlled rare-object interventions, and affordance-induced action revision.
    }
    \label{fig:benchmark-concept}
\end{figure}

For example, a plastic bag, a shopping cart, a fallen tree, and a road-closed sign may all appear as unusual objects in the ego lane, but they imply different driving responses. Some can be passed cautiously, some require slowing down and lateral adjustment, while others impose physical or normative constraints that require stopping or rerouting. The core challenge is therefore not merely \emph{what object is present}, but \emph{what action space the object affords}. We call this property \emph{decision-level driving affordance}. Figure~\ref{fig:benchmark-concept} highlights the key distinction between rare-object recognition and decision-level affordance prediction. The same nominal driving context may require different action revisions depending on whether the inserted object creates a physical obstruction, a position-sensitive conflict, a normative constraint, or a false-positive distraction. In this sense, \benchmark evaluates whether a model maps rare-object semantics to feasible meta-actions, rather than merely naming the object.

We formulate this problem as \emph{affordance-grounded meta-action prediction}. Given a front-view image, recent ego-motion history, and a navigation command, the model predicts a structured action pair consisting of one longitudinal action and one lateral action. This representation sits between object recognition and low-level control: it does not predict steering angles, throttle, or braking values, but instead captures high-level decisions such as \emph{slow down}, \emph{yield}, \emph{stop}, \emph{keep lane}, or \emph{nudge left}. Such decisions are shaped by spatial, physical, and normative constraints, making them a natural interface for evaluating whether a model understands the driving implication of a rare object.

Recent driving VLMs and VLAs have connected visual perception with language-level reasoning and action prediction, from driving-scene understanding~\citep{drivelm,drivevlm} to action-conditioned policies~\citep{lmdrive,autovla,alphadrive}. Meanwhile, existing benchmarks evaluate corner-case perception~\citep{coda}, corner-case VLM understanding~\citep{codalm}, driving generalization~\citep{bench2drive,navsim}, and action-level decisions~\citep{driveaction}. However, they do not isolate whether a rare object changes the feasible high-level action space. Since nominal driving logs are dominated by frequent behaviors, a model may learn common driving priors while still failing under rare-object interventions. This motivates a controlled diagnostic setting where the non-intervened driving context is fixed and only the rare object's decision-level affordance changes.

To this end, we introduce \benchmark, a counterfactual long-tail driving benchmark for decision-level affordance prediction. \benchmark contains 3,536 reviewed samples constructed from 29 base scenes, 50 obstacle types, spatial positions, and five affordance categories. Each sample provides a front-view image, ego-motion history, navigation command, and a set of acceptable longitudinal--lateral action pairs. Rare objects are inserted into otherwise fixed driving scenes while preserving road geometry, ego pose, navigation, and ego-motion context. This design turns accuracy into a diagnostic test of whether a model maps rare-object semantics to acceptable high-level driving actions. We use ``counterfactual'' in a controlled, diagnostic sense: the underlying driving scene is held fixed while the identity or position of an inserted object is varied, and the benchmark measures the corresponding change in the acceptable high-level action set. \benchmark does not model alternative trajectories, estimate causal effects in real traffic, or simulate closed-loop consequences. Accordingly, meta-action accuracy should not be interpreted as a measure of closed-loop driving safety.

We further study how to improve this capability in small VLMs. Large VLMs often contain stronger open-world knowledge but exceed the memory and compute budgets of in-vehicle platforms; we therefore study how to adapt small VLMs while preserving this knowledge. Yet direct fine-tuning on driving data can over-specialize small VLMs to frequent driving patterns and weaken the open-world knowledge needed for rare-object affordance reasoning~\citep{dea_blindspot,vlm_assisted_cl_vqa,deconfounded_lifelong_ad}. We therefore propose \method, a knowledge-preserving adaptation framework centered on a simple principle: adapt the model toward driving-specific action grounding without overwriting the open-world knowledge needed for rare-object reasoning. \method first uses a structured perception-to-decision interface to make the final longitudinal--lateral action explicit. It then constructs a conservative driving initialization through SLERP-based merging and trains \regmoe adapters on the frozen merged backbone. This design keeps the pretrained model as the dominant computation path while allowing driving regimes to activate different low-rank adaptation directions.

\looseness=-1 Our contributions are threefold. First, we formulate long-tail autonomous driving as decision-level driving affordance prediction, emphasizing the mapping from rare-object semantics to structured longitudinal--lateral actions. Second, we introduce \benchmark, a 3,536-sample counterfactual benchmark that diagnoses this capability through controlled rare-object interventions. Third, we propose \method, a knowledge-preserving adaptation recipe that combines SLERP-based conservative merging with regime-aware \regmoe adaptation to address the specialization--retention tension exposed by \benchmark; its contribution lies in the problem-driven integration and empirical analysis of these components for decision-level long-tail prediction, rather than in a new general MoE architecture.

\section{Related Work}
\label{sec:related}

\paragraph{Driving VLMs and VLAs.}
Recent driving VLMs connect visual observations, language instructions, reasoning traces, and driving decisions. DriveGPT4~\citep{drivegpt4}, DriveLM~\citep{drivelm}, DriveVLM~\citep{drivevlm}, and LMDrive~\citep{lmdrive} study language-supervised scene understanding, explanation, planning, and closed-loop driving. EMMA~\citep{emma}, Senna~\citep{senna}, OmniDrive~\citep{omnidrive}, and DriveMLM~\citep{drivemlm} further explore end-to-end multimodal driving policies. More recent reasoning- or action-oriented systems, such as DriveCoT~\citep{drivecot}, Reason2Drive~\citep{reason2drive}, DriveLMM-o1~\citep{drivelmmo1}, ReasonPlan~\citep{reasonplan}, Drive-R1~\citep{driver1}, AlphaDrive~\citep{alphadrive}, AutoVLA~\citep{autovla}, and Alpamayo-R1~\citep{alpamayor1}, introduce structured reasoning, reinforcement learning, or VLA-style action alignment. Recent evaluations also show that MLLMs can interpret individual driving frames while remaining unreliable in temporal dynamics, road-agent interactions, trajectory planning, and open-set reasoning~\citep{sreeram2025probing}. These works primarily ask whether a model can understand, explain, or execute a driving scene. Our focus is different: whether a rare object changes the feasible longitudinal--lateral action space under a controlled intervention.

\paragraph{Long-tail driving evaluation.}
Long-tail evaluation is central to autonomous driving because rare events often dominate safety risk. CODA~\citep{coda} studies corner-case object detection, CODA-LM~\citep{codalm} evaluates self-driving VLMs on corner cases, Bench2Drive~\citep{bench2drive} and NAVSIM~\citep{navsim} evaluate driving generalization, and DriveBench~\citep{drivebench} studies VLM reliability. Other efforts address safety-critical scene generation~\citep{chatscene}, object-level long-tail knowledge~\citep{token}, open VLA driving data~\citep{impromptu_vla}, realistic long-tail planning~\citep{interplan}, and action-level decision evaluation~\citep{driveaction}. Factorized OOD studies further show that aggregate robustness scores can obscure substantially different failure patterns across environmental shifts and their interactions~\citep{mallak2026robustness}. These benchmarks are valuable, but their evaluation units are mainly perception, VQA, planning, reliability, data coverage, or closed-loop behavior. \benchmark instead evaluates \emph{affordance-induced action revision}: the non-intervened driving context is held fixed, while an inserted rare object changes the acceptable longitudinal--lateral action set.

\paragraph{Affordance grounding and knowledge-preserving adaptation.}
Affordance grounding studies how observations imply feasible actions. SayCan~\citep{saycan} combines language reasoning with affordance functions, while RT-2~\citep{rt2}, OpenVLA~\citep{openvla}, and $\pi_0$~\citep{pi0} transfer vision-language knowledge to action prediction. In driving, VLA systems such as AutoVLA~\citep{autovla}, DriveMoE~\citep{drivemoe}, and DriveAction~\citep{driveaction} connect language-level reasoning with trajectories or action decisions. Our task differs in that the output is a structured high-level meta-action rather than a low-level motor command or continuous trajectory. Meanwhile, parameter-efficient fine-tuning methods such as LoRA~\citep{lora}, QLoRA~\citep{qlora}, AdaLoRA~\citep{adalora}, DoRA~\citep{dora}, PiSSA~\citep{pissa}, and VeRA~\citep{vera} reduce adaptation cost, but do not by themselves prevent knowledge loss. Forgetting remains a concern in model adaptation~\citep{loraforgets,mixearly,ewc,lwf}, including driving-specific VLM adaptation and continual learning~\citep{vlm_assisted_cl_vqa,dea_blindspot,deconfounded_lifelong_ad}. \method addresses this semantic-action tension by adapting a small VLM for driving-specific affordance prediction while preserving the open-world knowledge needed for rare-object reasoning.

\section{Task: Affordance-Grounded Meta-Action Prediction}
\label{sec:task}

Let $I$ denote a front-view driving image, $E$ denote the recent ego-motion history, and $n$ denote a navigation command. The model predicts a structured meta-action
\begin{equation}
    \mathbf{a} = (a_{\mathrm{lon}}, a_{\mathrm{lat}}),
\end{equation}
where $a_{\mathrm{lon}} \in \mathcal{A}_{\mathrm{lon}}$ is a longitudinal action and $a_{\mathrm{lat}} \in \mathcal{A}_{\mathrm{lat}}$ is a lateral action. The longitudinal action space contains high-level speed-control decisions such as \emph{keep speed}, \emph{slow down}, \emph{yield}, \emph{creep}, and \emph{stop}. The lateral action space contains high-level steering decisions such as \emph{keep lane}, \emph{nudge left or right}, and \emph{lane change left or right}.

This formulation avoids low-level control while targeting the semantic decision interface where rare-object affordances become actionable. For instance, rigid obstacles may require slowing and nudging, harmless distractors may allow keeping lane, and signs may impose normative constraints. The task therefore evaluates whether a model transforms object semantics, spatial layout, and context into a decision-level action pair. Meta-actions denote the immediate high-level response under the current observation rather than the terminal maneuver: a distant full blockage, for instance, is answered with deceleration on approach, and whether the vehicle ultimately comes to a complete stop is resolved by subsequent re-planning as the scene evolves, which is outside the scope of a single-frame decision.

For each sample $x=(I,E,n)$, the reference is not necessarily a single action but an acceptable set $\mathcal{P}(x)$ of longitudinal--lateral pairs. This reflects the fact that several high-level actions can be safe or semantically equivalent in a given scene. For example, a yield-or-stop situation may accept both \emph{yield, keep lane} and \emph{stop, keep lane}, while a left-biased obstacle may accept \emph{slow down, nudge right}. Unsafe directional choices are not included in $\mathcal{P}(x)$.

\begin{figure*}[t]
    \centering
    \includegraphics[width=0.98\linewidth]{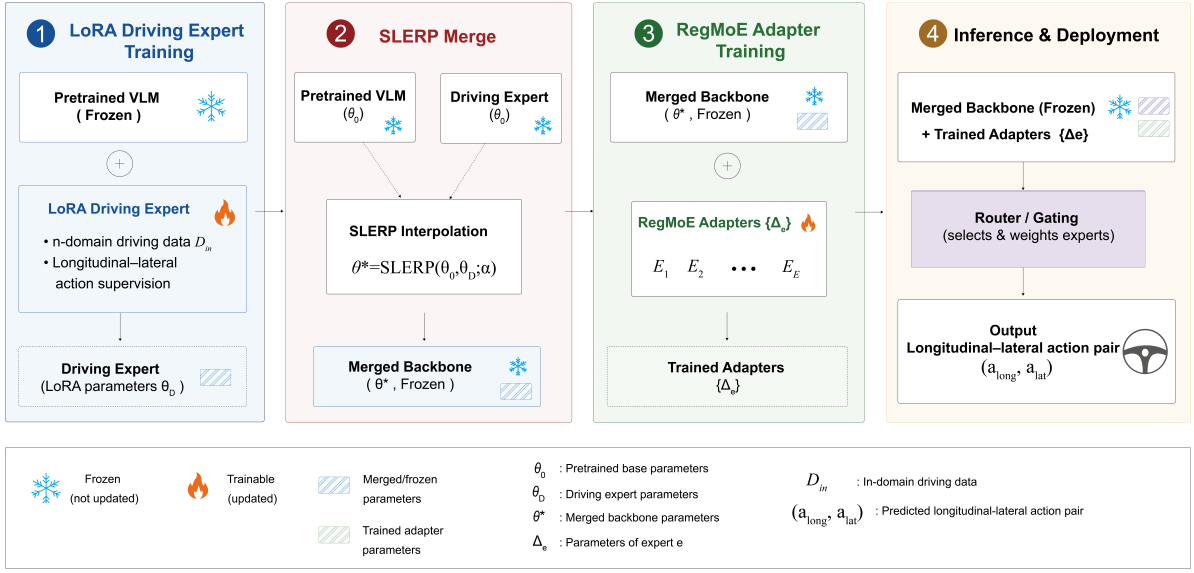}
    \caption{
    Overview of \method: SLERP constructs a conservative merged backbone, and \regmoe provides regime-aware low-rank adaptation on top of the frozen backbone.
    }
    \label{fig:kpa-pipeline}
\end{figure*}

\section{The \benchmark Benchmark}
\label{sec:benchmark}

\benchmark evaluates whether a driving VLM maps rare-object semantics to decision-level affordance, rather than merely recognizing the inserted object. Given a front-view image, ego-motion history, and navigation command, the model predicts an acceptable longitudinal--lateral meta-action. The benchmark contains 3,536 reviewed samples constructed from 29 base driving scenes.


We use the term \emph{counterfactual} to refer to controlled image-level interventions. For each base scene, we preserve the road geometry, ego-view perspective, ego-motion history, navigation command, and lead-vehicle context, while inserting rare objects at calibrated driving-relevant positions. Thus, \benchmark does not simulate full physical counterfactual trajectories or closed-loop world evolution. Instead, it isolates whether changing the rare object, while holding the non-intervened context fixed, changes the feasible high-level action space and whether the model predicts that change.

\paragraph{Affordance-oriented construction.}
Unlike recognition-oriented corner-case datasets, \benchmark groups rare objects by the action constraints they impose. It covers five affordance categories: living entities, nonliving entities, road hazards, full blockages, and false positives. These categories test different decision signatures, such as decelerating or yielding to living entities, slowing or nudging around rigid obstacles, decelerating toward a stop for full blockages, and avoiding overreaction to low-risk visual distractors. The goal is to evaluate affordance-induced action revision: visually similar objects may require different actions, while visually different objects may share the same feasible action set.

\paragraph{Counterfactual versions and labels.}
Each object-position intervention has two versions. The $v_{\mathrm{full}}$ version preserves surrounding traffic cues and evaluates affordance prediction under naturalistic context. The $v_{\mathrm{clean}}$ version removes moving vehicles and reduces traffic-context shortcuts, making the inserted object and road geometry more central to the decision. Both versions are included in the main benchmark because they represent complementary evaluation regimes rather than duplicate samples. Their breakdown is reported separately as a context-cue analysis.

\looseness=-1 Each sample is labeled with an acceptable set $\mathcal{P}(x)$ of longitudinal--lateral action pairs. Three reviewers independently annotate the 1,768 $v_{\mathrm{full}}$ images using a shared labeling guide, adjudicating disagreements into a consensus set; each $v_{\mathrm{clean}}$ label extends its adjudicated $v_{\mathrm{full}}$ counterpart with actions made safe by vehicle removal. Multiple safe actions are allowed when appropriate, while wrong or unsafe maneuvers are excluded. On the $v_{\mathrm{full}}$ split, the three annotators independently produce identical acceptable sets on 78.5\% of samples, with a set-valued Krippendorff's $\alpha$ (MASI distance) of 0.825; the remaining 21.5\% are resolved through pair-by-pair adjudication (Appendix~\ref{app:annotation-reliability}).

\paragraph{Validity checks.}
 Blind human ratings on 250 category-balanced synthetic images and 75 real controls show realism and plausibility (4.31/4.47 vs.\ 4.58/4.63 for real; 91.2\% of synthetic images with medians $\ge$4), and the cross-model ranking on a real control set correlates strongly with \benchmark (Spearman's $\rho=0.93$; Appendix~\ref{app:validity}).

\paragraph{Evaluation metric.}
We use pair accuracy as the primary metric:
\begin{equation}
    \mathrm{Acc}
    =
    \frac{1}{N}
    \sum_{i=1}^{N}
    \mathbf{1}
    \left[
    \hat{\mathbf{a}}_i \in \mathcal{P}(x_i)
    \right],
\end{equation}
where $\hat{\mathbf{a}}_i=(\hat a_{\mathrm{lon}},\hat a_{\mathrm{lat}})$ is the parsed prediction. We report overall pair accuracy and per-category accuracy. Because the non-intervened context is controlled, pair accuracy measures whether the model converts the inserted object's decision-level affordance into an acceptable structured action.

\section{Knowledge-Preserving Adaptation}
\label{sec:method}
\benchmark reveals that long-tail driving failure is not merely a perception failure, but a semantic-action grounding failure. This creates a specific adaptation challenge for small VLMs: the model must acquire driving-specific meta-action behavior while preserving the open-world object knowledge needed to reason about rare interventions. \method is designed around this semantic-action tension.

The framework contains three components, as shown in Figure~\ref{fig:kpa-pipeline}. First, a structured perception-to-decision interface makes the final longitudinal-lateral action explicit and parsable. Second, SLERP-based expert merging provides a conservative driving initialization that balances in-domain specialization and pretrained knowledge retention. Third, \regmoe introduces lightweight regime-aware specialization on top of the frozen merged backbone. Exploratory variants such as Fisher-guided adapters and GRPO-based optimization are discussed in Appendix~\ref{app:exploratory-adaptation}.

\subsection{Structured Perception-to-Decision Interface}
\label{sec:method-prompt}

Given $(I,E,n)$, the model is prompted to identify decision-relevant objects, infer spatial, physical, and normative constraints, and output a final action pair in a canonical format. The prompt is intentionally structured because free-form chain-of-thought can produce fluent but unparsable or action-inconsistent responses. This design is inspired by structured reasoning and chain-of-thought prompting in driving VLMs~\citep{drivecot,drivecot_pkrd,drivelmmo1,hmvlm,alpamayor1}, but we supervise only the final action tokens because \benchmark evaluates the decision interface rather than explanation fluency. During supervised adaptation, we optimize only the tokens corresponding to the final meta-action:
\begin{equation}
    \mathcal{L}_{\mathrm{sft}}(\theta)
    =
    -
    \sum_{t\in\mathcal{T}_{\mathbf{a}}}
    \log p_{\theta}(y_t \mid y_{<t}, I,E,n),
\end{equation}
where $\mathcal{T}_{\mathbf{a}}$ denotes the token positions of the final longitudinal and lateral decisions. This keeps supervision focused on the decision interface evaluated by \benchmark rather than on explanation style.

\subsection{SLERP-Based Expert Merging}
\label{sec:method-slerp}

Direct supervised fine-tuning can improve in-domain driving decisions, but it may also over-specialize the model toward frequent driving priors and weaken the open-world knowledge needed for rare-object affordance reasoning. We therefore decouple driving specialization from knowledge preservation. Instead of directly deploying a driving-finetuned model, we first obtain a driving expert and then interpolate it with the original pretrained model in weight space.

Let $\theta_0$ denote the pretrained small VLM. We train a driving expert with LoRA on the in-domain driving split and merge the LoRA update into the base weights to obtain a materialized driving expert $\theta_D$. This expert captures driving-specific action behavior, but may drift from the pretrained model's broader visual and commonsense representations. To balance specialization and retention, we construct the initialization $\theta^{\star}$ by interpolating $\theta_0$ and $\theta_D$:
\begin{equation}
    \theta^{\star}
    =
    \operatorname{SLERP}(\theta_0, \theta_D; \alpha),
\end{equation}
where $\alpha\in[0,1]$ controls the specialization--retention trade-off. A larger $\alpha$ moves the model closer to the driving expert, while a smaller $\alpha$ keeps it closer to the pretrained VLM.

This stage is related to the broader model-merging literature, which shows that weight-space combination can trade off task specialization and generalization~\citep{modelsoups,wisefit,taskarithmetic,ties,dare,adamerging}. We instantiate this idea with spherical linear interpolation (SLERP), which interpolates along the angular direction between two matched tensors rather than using linear averaging.

For each matched floating-point tensor, SLERP is computed as
\begin{equation}
\begin{aligned}
    \operatorname{SLERP}(u,v;\alpha)
    =
    \frac{\sin((1-\alpha)\Omega)}{\sin\Omega}u
    +
    \frac{\sin(\alpha\Omega)}{\sin\Omega}v,
\end{aligned}
\end{equation}
where $\Omega$ is the angle between the flattened normalized tensors. If either tensor has near-zero norm or $\sin\Omega$ is numerically unstable, we fall back to linear interpolation. Non-floating-point buffers are copied without interpolation.

  The resulting model $\theta^{\star}$ serves as a conservative driving initialization. It inherits part of the driving expert's action behavior while avoiding a full shift away from the pretrained VLM's open-world visual and commonsense knowledge. Adapter training is performed on top of this frozen merged backbone.

\subsection{Regime-Aware LoRA Mixture-of-Experts}
\label{sec:method-regmoe}

The SLERP-merged model $\theta^\star$ provides a conservative driving initialization, but it still represents all driving situations with a shared adaptation direction. This is limiting for long-tail affordance prediction, because routine following, intentional maneuvers, and safety-reactive behaviors often require different decision priors. We therefore freeze $\theta^\star$ and train \regmoe, a regime-aware LoRA mixture-of-experts module, as the final adaptation stage. Our design follows the broader line of MoE-style parameter-efficient adaptation, where multiple low-rank experts are used to increase specialization without updating the full backbone~\citep{loramoe,mole,moelora,mixlora}. Unlike these general-purpose adapter mixtures, \regmoe uses driving decision regimes as an explicit routing signal.

For a frozen projection $W_0$, \regmoe attaches a set of LoRA experts $\{\Delta_e\}_{e=1}^{E}$, where each expert represents one low-rank correction direction. Given a hidden token representation $x$ and a coarse driving regime $t$, the router predicts an expert mixture by combining token-level evidence with a regime-conditioned bias:
\begin{equation}
\label{eq:regmoe-routing}
    \boldsymbol{\pi}(x,t)
    =
    \operatorname{softmax}
    \bigl(g(x)+b_t\bigr).
\end{equation}
The adapted projection is computed as
\begin{equation}
\label{eq:regmoe-output}
    y
    =
    W_0x
    +
    \frac{\alpha}{r}
    \sum_{e=1}^{E}
    \pi_e(x,t)\Delta_e(x).
\end{equation}
This keeps the frozen SLERP projection as the main computation path, while allowing different driving regimes to activate different low-rank adaptation directions. This regime-conditioned routing is related to domain- or distribution-specialized expert routing~\citep{lifelongmoe,modula} and recent MoE designs for multimodal or driving models~\citep{moellava,omnismola,drivemoe}. Unlike DriveMoE~\citep{drivemoe}, which routes full skill-specialized experts in a trajectory decoder by semantic scenario category, \regmoe routes low-rank residual experts over a frozen SLERP-merged backbone using decision-level action labels, targeting structured meta-action prediction rather than trajectory generation; we do not claim the supervised-routing principle as novel.

\paragraph{Adaptation-time and inference-time routing.}
We derive $t$ from the supervised driving action label during adaptation and group samples into three coarse regimes: Routine for keep-lane or keep-speed behavior, Maneuver for turning or lane-changing behavior, and Reactive for nudge, stop, decelerate, yield, or speed-adjustment behavior. Regime ids are derived from the fine-grained 72-label action ontology of the in-domain training data; for example, \emph{turn left}, \emph{turn right}, \emph{u-turn}, and lane-change labels are grouped into Maneuver. The regimes supervise routing only and do not introduce additional output tokens; the evaluated action space is unchanged. The regime label is an adaptation-time signal only: the bias $b_t$ is added to the gate logits only when a regime id is supplied to the MoE layer during training. At inference, no regime id is available, the bias branch is skipped, and the expert mixture is determined exclusively from the input hidden states:
\begin{equation}
\label{eq:regmoe-routing-inference}
    \boldsymbol{\pi}(x)
    =
    \operatorname{softmax}\bigl(g(x)\bigr).
\end{equation}
All reported \method results use this regime-free inference path: no ground-truth action, regime label, or forced expert assignment is supplied at test time. This grouping is intentionally coarse: it does not define a driving taxonomy, but provides a lightweight decision-regime signal for expert routing.

The adapters are trained with weighted action-token supervision and routing regularization:
\begin{equation}
\label{eq:regmoe-loss}
    \mathcal{L}_{\mathrm{RegMoE}}
    =
    \lambda_{\mathrm{ans}}\mathcal{L}_{\mathrm{CE}}
    +
    \lambda_{\mathrm{lb}}\mathcal{L}_{\mathrm{lb}}
    -
    \lambda_{\mathrm{sep}}\mathcal{L}_{\mathrm{sep}} .
\end{equation}
Here $\mathcal{L}_{\mathrm{CE}}$ supervises the final longitudinal--lateral action tokens, $\mathcal{L}_{\mathrm{lb}}$ prevents expert collapse by balancing expert usage, and $\mathcal{L}_{\mathrm{sep}}$ encourages different driving regimes to form distinct expert mixtures. Implementation details are in Appendix~\ref{app:implementation}.

\subsection{Training Procedure}
\label{sec:method-training}

We train \method in two stages. First, we train a LoRA driving expert on the in-domain split $\mathcal{D}_{\mathrm{in}}$, materialize the update into the base model to obtain $\theta_D$, and merge it with the pretrained model $\theta_0$ using SLERP to obtain $\theta^\star$. Second, we freeze $\theta^\star$ and train the \regmoe parameters $\psi$, including LoRA expert matrices, token router, and regime-conditioned routing biases, with the loss in Eq.~\ref{eq:regmoe-loss}. The final model consists of the frozen SLERP-merged backbone and the trained \regmoe adapters. No samples from \benchmark are used for training.

\section{Experiments}
\label{sec:experiments}

We evaluate \method on nominal driving samples and controlled rare-object interventions in \benchmark. The evaluation covers overall and category-level pair accuracy, $v_{\mathrm{full}}$/$v_{\mathrm{clean}}$ context-cue analysis, language-side knowledge retention, and ablations, testing whether \method improves long-tail affordance prediction while preserving the pretrained knowledge needed for rare-object reasoning.

\subsection{Experimental Setup}

\paragraph{Models.}
 Our main backbone is Qwen3-VL-2B~\citep{qwen3vl}, a deployable small VLM. We compare the pretrained model, LoRA supervised fine-tuning, SLERP-based merging, and the final \method model with \regmoe adapters. Exploratory Fisher and GRPO variants are in Appendix~\ref{app:exploratory-adaptation}.

\paragraph{Data.}
The in-domain driving data comes from a multi-camera driving-log corpus with 10{,}000 clips across 26 ODD categories. We perform a 90/10 clip-level multi-label stratified split, producing 9{,}002 training clips with 199{,}802 samples and 998 held-out test clips with 21{,}857 samples. The clip-level split prevents frames from the same recording from appearing in both training and evaluation.
For nominal driving evaluation, we use a 3{,}613-sample held-out subset drawn from the test clips. It covers all 998 test clips, all 26 ODD categories, and 72 action labels, while matching the full test distribution closely, with at most 0.24\% per-action and 0.68\% per-ODD deviation. \benchmark contains 3{,}536 counterfactual samples with acceptable action-pair sets.

\paragraph{Parsing and judging.}
All models use the same structured perception-to-decision prompt and greedy decoding strategy, and are scored with an identical three-stage pipeline. First, a deterministic rule-based parser isolates the model's committed decision span from the raw response; outputs with no unique decision span are marked invalid. Second, a text-only LLM decision normalizer (DeepSeek-v4-Pro, temperature~$0$) reads only the extracted text---not the image or reference labels---and maps it to the canonical longitudinal--lateral action pair, so that semantically equivalent non-canonical phrasings are normalized consistently. Third, the normalized pair is scored deterministically against the acceptable set $\mathcal{P}(x)$; the LLM does not determine correctness. Invalid outputs are counted as incorrect. The full pipeline and per-model invalid-output rates are in Appendix~\ref{app:prompting-parsing}.

\paragraph{Metrics.}
We report pair accuracy on the in-domain split and \benchmark, with per-category, $v_{\mathrm{full}}$/$v_{\mathrm{clean}}$, and language-side retention accuracy.

\vspace{-0.4em}

\begin{table}[t]
\centering
\footnotesize
\setlength{\tabcolsep}{3pt}
\renewcommand{\arraystretch}{1.00}
\begin{tabularx}{\linewidth}{@{}>{\raggedright\arraybackslash}X rr@{}}
\toprule
Model & \shortstack{In-domain\\Acc.} & \shortstack{\benchmark\\Acc.} \\
\midrule 
Curious-VLA-3B~\citep{chen2026devil}      &  5.9 &  1.9 \\
Cosmos-Reason2-8B~\citep{nvidia2025cosmosreason2}   & 25.0 & 37.1 \\
Qwen3-VL-4B~\citep{qwen3vl}        & 28.4 & 56.9 \\
Qwen3-VL-8B~\citep{qwen3vl}        & 29.1 & 65.5 \\
Qwen3-VL-32B~\citep{qwen3vl}       & 32.2 & 59.0 \\
Alpamayo-1.5-10B~\citep{alpamayor1}        & 32.1 & 59.3 \\
AutoDrive-R$^2$-7B~\citep{yuan2025autodrive}       & 32.2 & 48.6  \\
GPT-5.5~\citep{openai2026gpt55}             & \textbf{40.0} & \textbf{83.0} \\
\midrule
Cosmos-Reason2-2B~\citep{nvidia2025cosmosreason2}   &  9.4 & 34.9 \\
Qwen3-VL-2B        & 12.4 & 50.3 \\
Qwen3-VL-2B + LoRA-SFT & \textbf{58.3} & 32.4 \\
Qwen3-VL-2B + SLERP    & 55.9 & 53.0 \\
Qwen3-VL-2B + \method  & 52.8 & \textbf{60.8} \\
\bottomrule
\end{tabularx}
\caption{Main results. Accuracy is computed over parsed longitudinal--lateral action pairs. All Qwen3-VL-2B variants use the same backbone; larger models are evaluated without task-specific fine-tuning.}
\label{tab:main-results}
\end{table}

\subsection{Main Results}

\begin{table*}[t]
\centering
\small
\renewcommand{\arraystretch}{1.08}
\begin{tabularx}{\linewidth}{Yccccc}
\toprule
Model & Living entity & Nonliving entity & Road hazard & Full block & False positive \\
\midrule
\multicolumn{6}{l}{\emph{Reference models}} \\
GPT-5.5 & 91.7 & 64.8 & 79.3 & 81.0 & 91.5 \\
Alpamayo-1.5-10B & 99.7 & 13.8 & 27.6 & 66.6 & 85.8 \\
AutoDrive-R$^2$-7B & 39.7 & 25.9 & 29.6 & 53.8 & 83.5 \\
Qwen3-VL-32B & 45.7 & 45.8 & 44.5 & 54.5 & 82.8 \\
Qwen3-VL-8B & 70.7 & 49.6 & 51.1 & 64.8 & 84.4 \\
Qwen3-VL-4B & 42.0 & 46.4 & 43.1 & 48.3 & 89.0 \\
Cosmos-Reason2-8B & 50.3 & 16.1 & 21.0 & 46.9 & 65.3 \\
Cosmos-Reason2-2B & 35.9 & 11.2 & 13.5 & 34.1 & 73.3 \\
Curious-VLA-3B & 0.29 & 0.86 & 0.29 & 0.00 & 3.68 \\
\midrule
\multicolumn{6}{l}{\emph{Qwen3-VL-2B variants}} \\
Base & 36.2 & 30.0 & 35.6 & 38.3 & 91.0 \\
+ LoRA-SFT & 29.3 & 11.2 & 10.9 & 16.9 & 88.3 \\
+ SLERP & 44.0 & 33.7 & 35.1 & 41.0 & 88.0 \\
+ \method & 82.5 & 22.2 & 37.9 & 69.7 & 83.7 \\
\bottomrule
\end{tabularx}
\caption{Per-category pair accuracy on the $v_{\mathrm{full}}$ split of \benchmark. The breakdown evaluates whether models predict appropriate action pairs across different affordance categories under naturalistic surrounding context.}
\label{tab:category-results}
\end{table*}

Table~\ref{tab:main-results} shows a clear tension between nominal driving adaptation and counterfactual long-tail robustness. LoRA SFT improves in-domain accuracy from 12.4\% to 58.3\%, but reduces \benchmark accuracy from 50.3\% to 32.4\%, indicating over-specialization to frequent driving priors. SLERP mitigates this degradation, and the full \method further improves \benchmark accuracy to 60.8\%, outperforming the pretrained 2B backbone, LoRA SFT, and SLERP by 10.5, 28.4, and 7.8 points, respectively. Among reference models, GPT-5.5 and Qwen3-VL-8B remain stronger, while driving-specialized models such as Alpamayo-1.5-10B and AutoDrive-R$^2$-7B do not consistently outperform general VLMs. This suggests that existing driving experts may learn trajectory prediction, planning priors, or driving-domain instruction following, but still lack robust rare-object affordance grounding under our structured action-pair evaluation. Overall, \benchmark exposes a capability gap that is not explained by model scale, nominal driving adaptation, or driving-specific pretraining alone.

Curious-VLA-3B is a trajectory-specialized VLA reference: 95.6\% of its outputs are invalid under the structured decision interface (Appendix~\ref{app:prompting-parsing}), so its score reflects interface incompatibility rather than driving reasoning, and it is not used as evidence for \method's advantage.

\subsection{Affordance Category Analysis}

Table~\ref{tab:category-results} shows that \method mainly improves safety-critical categories. Compared with SLERP, it raises living-entity accuracy from 44.0\% to 82.5\% and full-block accuracy from 41.0\% to 69.7\%, suggesting that regime-aware adaptation better activates the cautious-deceleration responses required by safety-critical interventions. However, the gains are not uniform: \method remains weaker on nonliving entities and false positives, indicating possible overreaction to passable rigid objects or low-risk distractors. Reference models show a similar issue. Alpamayo-1.5-10B and AutoDrive-R$^2$-7B perform well on some coarse cases, such as living entities, full blocks, or false positives, but struggle on finer affordance categories such as nonliving obstacles and road hazards. This suggests that driving-specialized models may learn broad driving priors, but still lack fine-grained rare-object affordance grounding.

The gains are not a globally conservative shift: on nominal driving, \method keeps speed on 48.3\% of samples and its stop rate (13.8\%) matches the ground truth (13.6\%), while a constant \emph{slow down, keep lane} policy would score only 24.7\% versus \method's 52.8\%. On \benchmark, predictions shift toward \emph{slow down} (84.7\%) rather than \emph{stop} or \emph{yield} (0.7\% each), indicating an intervention-conditioned response (Appendix~\ref{app:action-distribution}); the nonliving and false-positive regressions stem mainly from lateral errors on passable objects (Appendix~\ref{app:error-analysis}).

\subsection{Ablation Study}


Table~\ref{tab:ablation} isolates the contribution of each component.
The ablation results show that the structured prompt is the most important component: replacing it with a direct action query reduces accuracy by 16.29 points. Removing SLERP initialization reduces accuracy by 4.98 points, confirming that the merged backbone provides a better starting point for long-tail affordance prediction. Replacing \regmoe with a single LoRA adapter lowers accuracy by 3.00 points, while removing the regime-conditioned routing bias lowers accuracy by 4.75 points. Together, these ablations indicate that both the initialization and the routing signal contribute to the final counterfactual performance.

Across four seeds, \regmoe ($59.72\pm0.88$ on $v_{\mathrm{full}}$) also outperforms a parameter-matched rank-48 single LoRA ($55.52\pm0.99$) in every run (paired $+4.20\pm0.58$; Appendix~\ref{app:seed-variance}), unlike the non-matched single-LoRA ablation above.

\begin{table}[t]
\centering
\small
\renewcommand{\arraystretch}{1.08}
\begin{tabularx}{\linewidth}{Ycc}
\toprule
Variant & $v_{\mathrm{full}}$ Acc. (\%) & Drop \\
\midrule
Full \method & 60.07 & -- \\
w/o structured prompt & 43.78 & -16.29 \\
w/o SLERP initialization & 55.09 & -4.98 \\
Single LoRA adapter & 57.07 & -3.00 \\
\regmoe w/o regime bias & 55.32 & -4.75 \\
\bottomrule
\end{tabularx}
\caption{Ablation study on the $v_{\mathrm{full}}$ split of \benchmark, with drops computed relative to Full \method.}
\label{tab:ablation}
\end{table}

\subsection{Context-Cue Analysis}

Table~\ref{tab:vfull-vclean} compares $v_{\mathrm{full}}$ and $v_{\mathrm{clean}}$, where the gap is defined as $v_{\mathrm{clean}}-v_{\mathrm{full}}$. Most general VLMs improve slightly after background vehicles are removed, suggesting that surrounding traffic cues can sometimes distract from the inserted object's affordance. In contrast, LoRA SFT drops from 34.6\% to 30.2\%, indicating stronger reliance on nominal context shortcuts learned from in-domain driving data. \method shows only a small gap of +1.36, close to GPT-5.5 and smaller than SLERP, suggesting more stable grounding across both context conditions. Notably, Cosmos-Reason2 models degrade in the clean setting, which further shows that driving or spatial-reasoning specialization does not guarantee robustness when contextual cues are reduced. Overall, the paired $v_{\mathrm{full}}$/$v_{\mathrm{clean}}$ design reveals whether models ground decisions in the inserted object's affordance or rely on surrounding scene context.

\begin{table}[t]
\centering
\small
\renewcommand{\arraystretch}{1.08}
\begin{tabularx}{\linewidth}{Yccc}
\toprule
Model & $v_{\mathrm{full}}$ & $v_{\mathrm{clean}}$ & Gap \\
\midrule
\multicolumn{4}{l}{\emph{Reference models}} \\
GPT-5.5 & 82.18 & 83.77 & +1.59 \\
Alpamayo-1.5-10B & 59.79 & 58.71 & -1.08 \\
AutoDrive-R$^2$-7B & 48.08 & 49.21 & +1.13 \\
Qwen3-VL-32B & 56.05 & 61.93 & +5.88 \\
Qwen3-VL-8B & 65.10 & 65.84 & +0.74 \\
Qwen3-VL-4B & 55.66 & 58.14 & +2.48 \\
Cosmos-Reason2-8B & 40.95 & 33.31 & -7.64 \\
Cosmos-Reason2-2B & 35.58 & 34.22 & -1.36 \\
Curious-VLA-3B & 1.19 & 2.55 & +1.36 \\
\midrule
\multicolumn{4}{l}{\emph{Qwen3-VL-2B variants}} \\
Base & 48.70 & 51.98 & +3.28 \\
+ LoRA-SFT & 34.62 & 30.15 & -4.47 \\
+ SLERP & 50.57 & 55.43 & +4.86 \\
+ \method & 60.07 & 61.43 & +1.36 \\
\bottomrule
\end{tabularx}
\caption{Context-cue analysis on \benchmark. }
\label{tab:vfull-vclean}
\end{table}

\paragraph{Robustness and efficiency checks.}
On 75 naturally occurring corner-case images from unseen scenes, the same-backbone ordering is preserved (\method 61.3\% $>$ SLERP 52.0\% $>$ pretrained 49.3\% $>$ LoRA SFT 30.7\%). Removing all nine Alpamayo-sourced scenes leaves \method's gains intact ($+11.6$/$+31.0$/$+8.0$ over the pretrained model, LoRA SFT, and SLERP), and Alpamayo-1.5-10B shows no home-source advantage. \regmoe adds 19.8M parameters (0.93\%) and $\approx$0.2~GB peak memory, recovering 69\% of the 2B-to-8B gap within the 2B memory class (Appendices~\ref{app:validity}--\ref{app:efficiency}).

\section{Conclusion}
\label{sec:conclusion}

We studied long-tail autonomous driving as \emph{decision-level driving affordance}: beyond recognizing rare objects, models must infer how they change feasible longitudinal and lateral actions. For evaluation, we introduced \benchmark, a controlled counterfactual benchmark that isolates affordance-induced action revision in fixed driving scenes.
Our experiments reveal a tension between nominal driving adaptation and long-tail robustness: LoRA SFT improves in-domain accuracy yet degrades \benchmark performance. In response, \method combines structured perception-to-decision prompting, SLERP-based conservative merging, and regime-aware LoRA experts to improve rare-object affordance prediction while preserving open-world knowledge. These results suggest that robust long-tail driving requires grounding rare-object semantics in structured action affordances.

\newpage
\section{Limitations}
This paper has several limitations. First, \benchmark is a controlled image-level diagnostic benchmark, not a full physical counterfactual or closed-loop driving evaluation. It measures whether models map rare-object semantics to high-level longitudinal-lateral actions, but does not simulate object dynamics, ego-action feedback, sensor noise, or downstream planner-control interactions.
Second, our task uses abstract meta-actions rather than continuous trajectories or low-level control. This helps isolate decision-level affordance prediction, but does not evaluate trajectory feasibility, vehicle kinematics, comfort, or multi-agent interaction. The benchmark also relies on generated counterfactual images and acceptable action-pair labels, which may contain visual artifacts, imperfect object placement, or residual subjectivity despite review.
Finally, our adaptation study focuses on a small VLM setting. Although \method improves long-tail affordance prediction for Qwen3-VL-2B, its gains are not uniform across affordance categories, and the current regime routing remains coarse. Future work should evaluate more backbones, real long-tail logs, finer-grained affordance routing, and closed-loop deployment settings.

Finally, improved meta-action accuracy on \benchmark should not be interpreted as evidence of real-world safety. Overreaction to harmless objects may cause unnecessary braking, and incorrect lateral avoidance may introduce new hazards; generated-image artifacts and the limited geographic, environmental, and traffic diversity of the benchmark may further produce uneven generalization across road environments, lighting, weather, countries, and vulnerable road users. \benchmark is a diagnostic stress test and is not a substitute for trajectory-level simulation, closed-loop evaluation, and platform-specific validation.


\bibliography{custom}

\clearpage

\appendix

\section{Benchmark Construction Details}
\label{app:benchmark-details}

\paragraph{Construction overview.}
\benchmark is constructed through a five-stage pipeline: (1)~base scene selection, (2)~obstacle taxonomy design, (3)~counterfactual image generation, (4)~paired $v_{\mathrm{full}}$/$v_{\mathrm{clean}}$ construction, and (5)~acceptable action-pair labeling with multi-round quality assurance. The pipeline keeps the non-intervened driving context fixed while varying the rare object and its spatial placement, so that evaluation focuses on affordance-induced action changes rather than generic scene understanding. The final benchmark contains 3,536 reviewed samples constructed from 29 base driving scenes and 50 obstacle types spanning five affordance categories. Each sample provides a front-view image, ego-motion history, navigation command, and a set of acceptable longitudinal--lateral action pairs. The complete dataset statistics are reported in Table~\ref{tab:benchmark-stats}.

\begin{figure}[h]
    \centering
    \includegraphics[width=\linewidth]{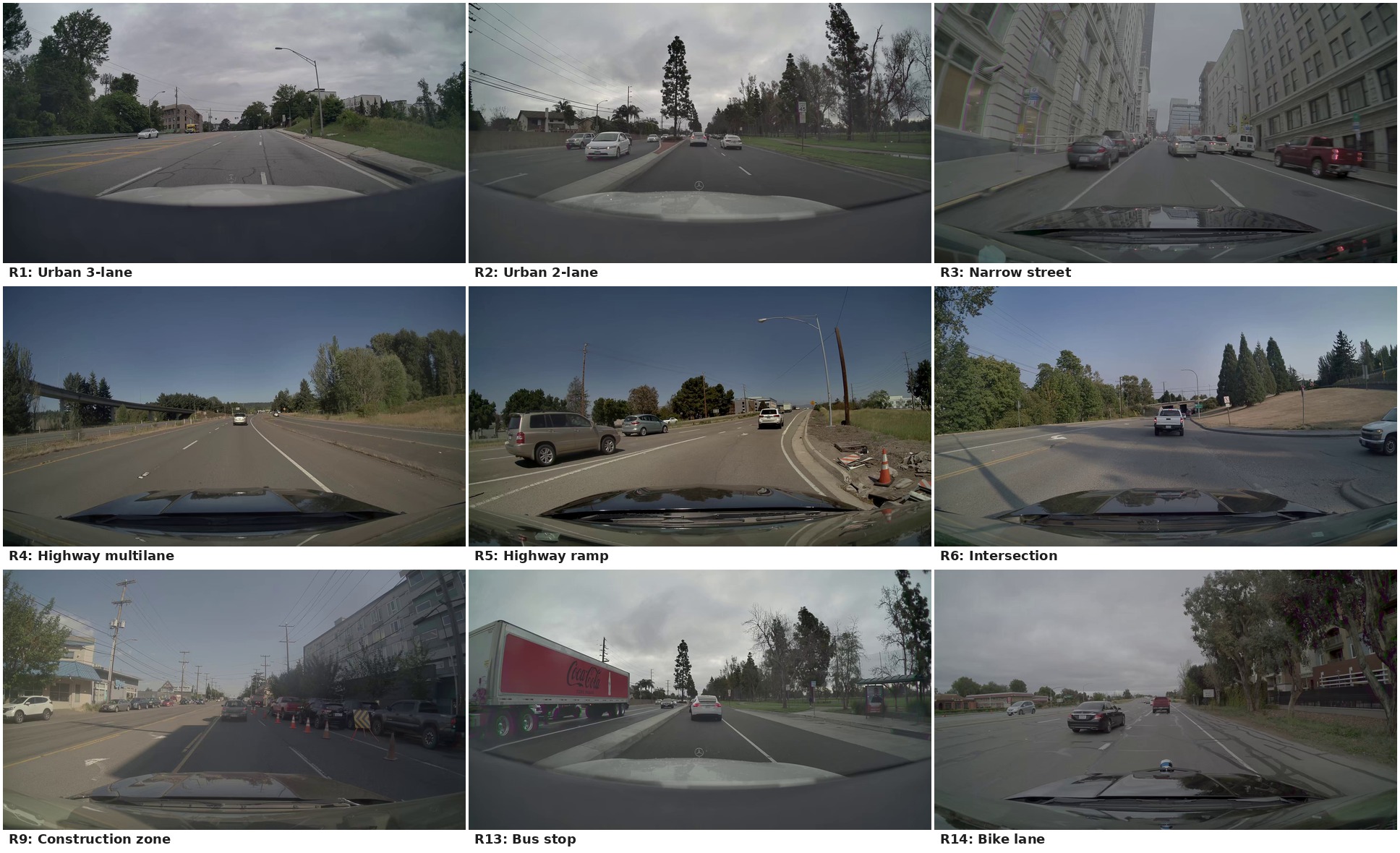}
    \caption{The 9 base scenes from the released dataset of Alpamayo-R1, covering urban, highway, intersection, construction, and bike-lane scenarios.}
    \label{fig:rbase-grid}
\end{figure}

\begin{figure}[h]
    \centering
    \includegraphics[width=\linewidth]{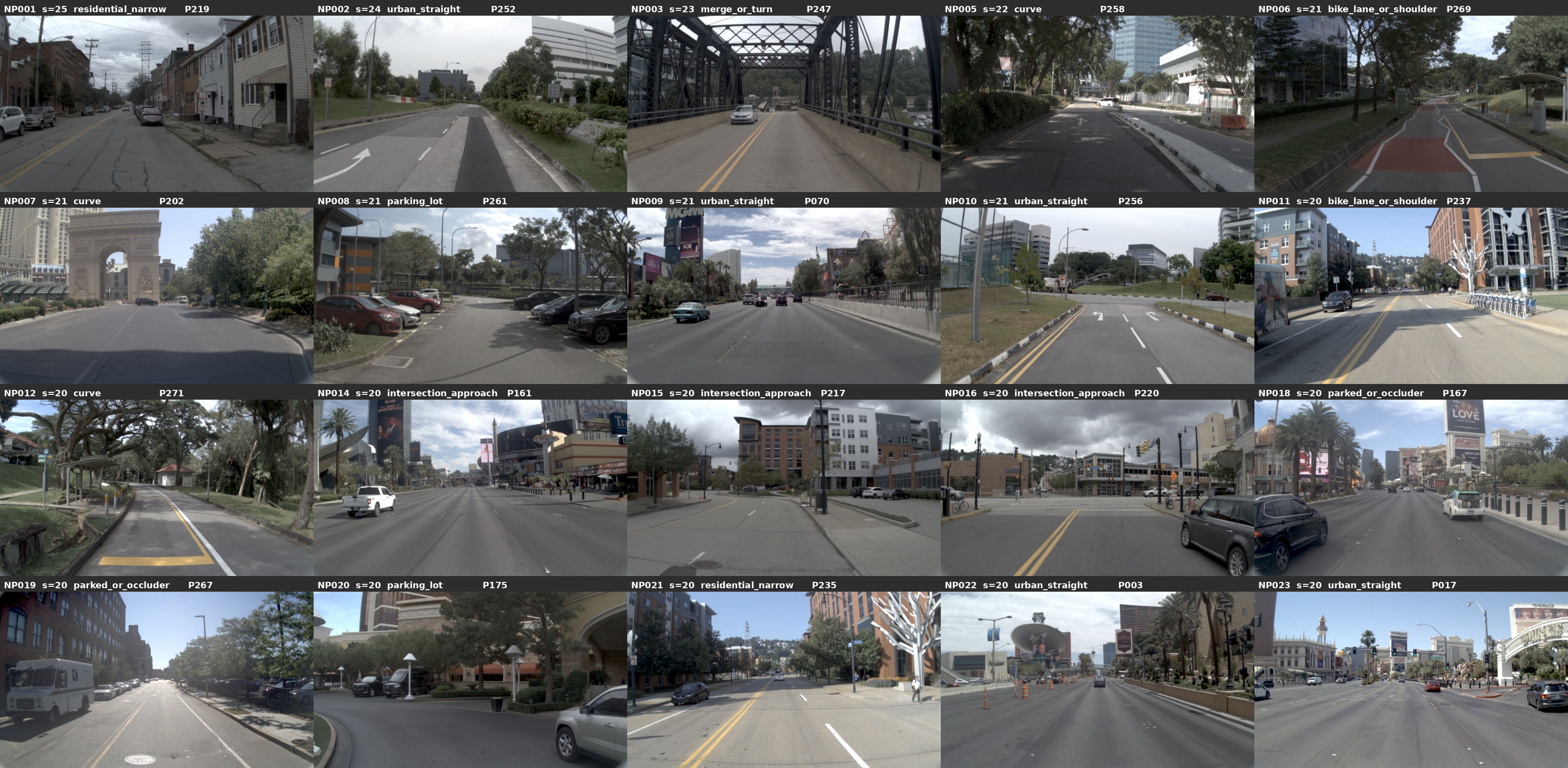}
    \caption{The 20 base scenes selected from nuPlan mini, extending coverage to residential corridors, curved roads, merge zones, occlusion-heavy streets, and parking areas.}
    \label{fig:npbase-grid}
\end{figure}

\paragraph{Base scene selection.}
The 29 base scenes come from two sources: 9 base scenes from the released dataset of Alpamayo-R1~\citep{alpamayor1} and 20 base scenes selected from nuPlan mini~\citep{nuplan}. Each base scene is selected to satisfy the following requirements: (i)~the front-facing camera provides a clear view of the ego lane and road ahead; (ii)~lane boundaries, road edges, or drivable regions are visually identifiable; (iii)~an insertion zone of approximately 15--35 meters ahead is available and not fully occluded by a lead vehicle; (iv)~the correct driving action is decidable by a human given the scene geometry and the inserted object; (v)~no traffic light, stop sign, or dense traffic queue already dominates the action decision; and (vi)~the image has adequate quality without severe blur, glare, or adverse weather. These criteria make the inserted object the primary source of decision change. The Alpamayo base scenes cover urban multi-lane roads, narrow residential streets, highway segments, intersections, construction zones, bus stops, and bike-lane corridors (Figure~\ref{fig:rbase-grid}). The nuPlan base scenes extend coverage to curved roads, merge zones, occlusion-heavy streets, and parking areas (Figure~\ref{fig:npbase-grid}). Each base scene declares which insertion types it supports (e.g., center obstacle, left-biased, full-lane blockage) and which it does not, so that obstacle placement respects scene geometry.

\begin{figure*}[t]
    \centering
    \includegraphics[width=0.95\linewidth]{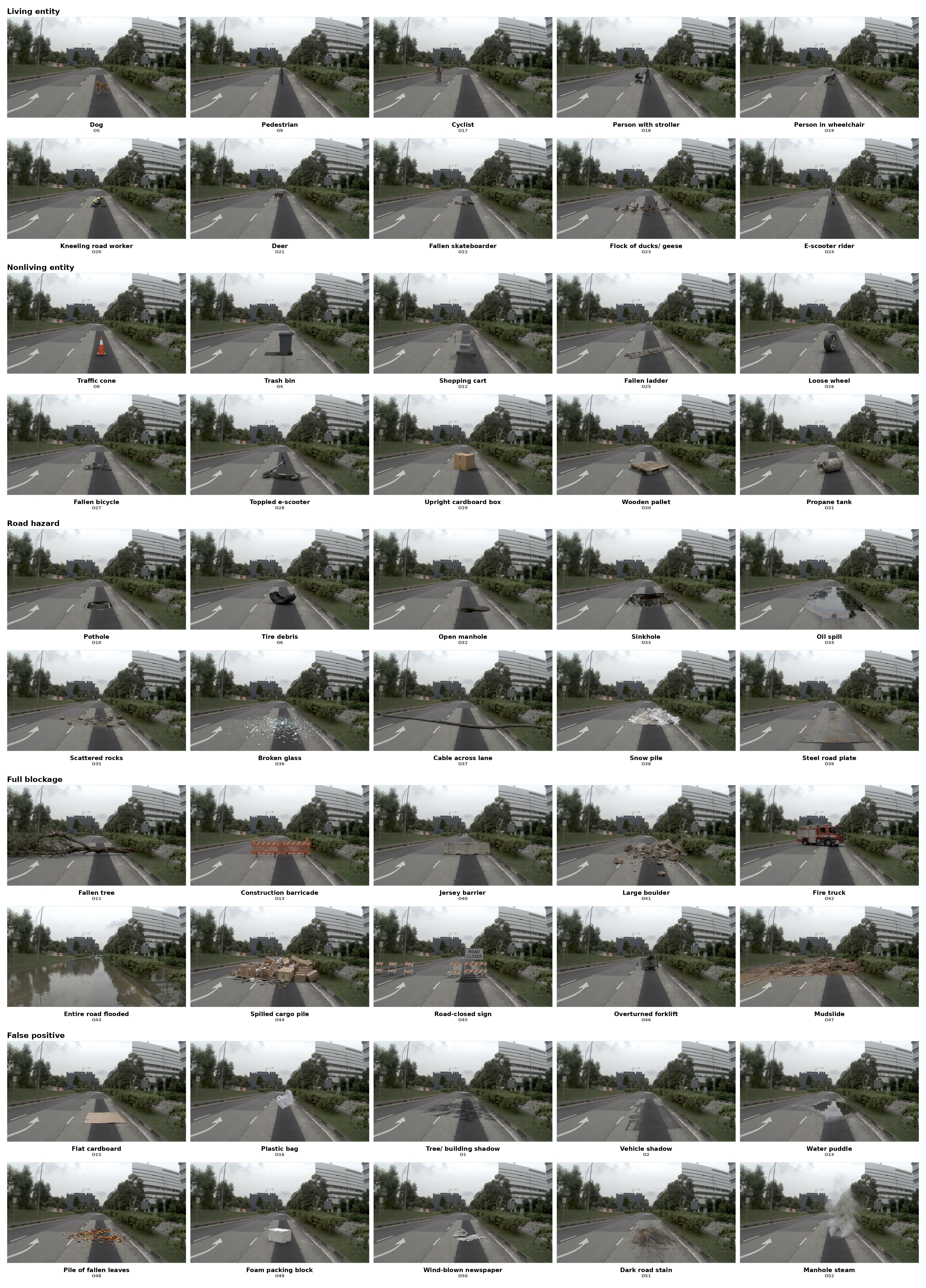}
    \caption{Gallery of all 50 obstacle types inserted into the same nuPlan base scene, organized by affordance category.}
    \label{fig:obstacle-gallery}
\end{figure*}

\paragraph{Obstacle taxonomy.}
The benchmark uses 50 obstacle types organized into five affordance categories. Unlike recognition-oriented benchmarks that group objects by visual appearance, our taxonomy groups objects by the action constraints they impose on the ego vehicle. Table~\ref{tab:obstacle-taxonomy} lists all 50 types. Figure~\ref{fig:obstacle-gallery} shows all 50 types inserted into the same base scene.  \emph{Living entities} (10~types) include pedestrians, cyclists, animals, and other vulnerable road users that typically require cautious deceleration, yielding, or stopping. \emph{Nonliving entities} (10~types) include rigid fallen or displaced objects such as traffic cones, ladders, and shopping carts that typically require slowing down and lateral adjustment. \emph{Road hazards} (10~types) include surface-level dangers such as potholes, oil spills, and open manholes that constrain speed and lateral path. \emph{Full blockages} (10~types) include large obstructions such as fallen trees, Jersey barriers, and fire trucks that obstruct the ego lane and require the vehicle to decelerate and ultimately stop or reroute. \emph{False positives} (10~types) include visually salient but physically harmless objects such as flat cardboard, shadows, and plastic bags that test whether models avoid overreacting to low-risk visual distractors.

\clearpage

\begin{table}[h]
\centering
\small
\renewcommand{\arraystretch}{1.05}
\begin{tabularx}{\linewidth}{lY}
\toprule
Category & Obstacle types \\
\midrule
Living entity &
Dog, Pedestrian, Cyclist, Person with stroller, Person in wheelchair, Kneeling road worker, Deer, Fallen skateboarder, Flock of ducks/geese, E-scooter rider \\
\midrule
Nonliving entity &
Traffic cone, Trash bin, Shopping cart, Fallen ladder, Loose wheel, Fallen bicycle, Toppled e-scooter, Upright cardboard box, Wooden pallet, Propane tank \\
\midrule
Road hazard &
Pothole, Tire debris, Open manhole, Sinkhole, Oil spill, Scattered rocks, Broken glass, Cable across lane, Snow pile, Steel road plate \\
\midrule
Full blockage &
Fallen tree, Construction barricade, Jersey barrier, Large boulder, Fire truck (blocking), Entire road flooded, Spilled cargo pile, Road-closed sign, Overturned forklift, Mudslide \\
\midrule
False positive &
Flat cardboard, Plastic bag, Tree/building shadow, Vehicle shadow, Water puddle, Pile of fallen leaves, Foam packing block, Wind-blown newspaper, Dark road stain, Manhole steam \\
\bottomrule
\end{tabularx}
\caption{The 50 obstacle types in \benchmark, organized by affordance category. Each category is defined by the action constraints it imposes rather than by visual appearance.}
\label{tab:obstacle-taxonomy}
\end{table}

\begin{figure*}[t]
    \centering
    \includegraphics[width=0.95\linewidth]{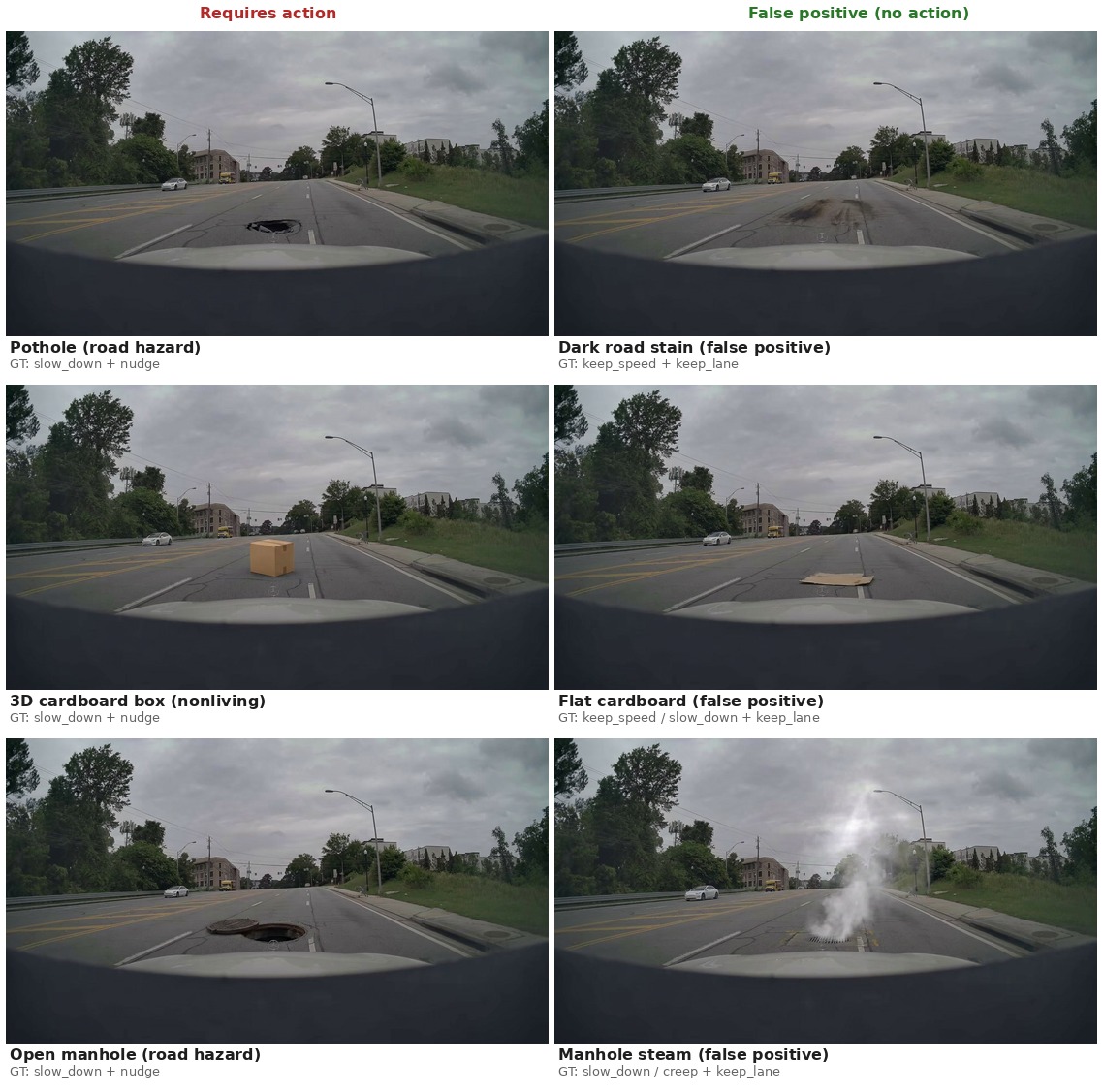}
    \caption{Diagnostic contrast pairs. Each row shows two visually similar objects inserted into the same base scene. The left column requires action revision (slow down and nudge); the right column is a false positive (keep speed or keep lane). The same road geometry and context are preserved, so only the object's affordance differs.}
    \label{fig:diagnostic-contrasts}
\end{figure*}

\paragraph{Diagnostic contrast design.}
A key design choice in \benchmark is the inclusion of \emph{diagnostic contrast pairs}: pairs of visually similar objects that belong to different affordance categories and therefore require different driving actions. For example, a pothole (road hazard: slow down and nudge) and a dark road stain (false positive: keep speed and keep lane) both appear as dark patches on the road surface but have opposite affordance implications. Similarly, an upright cardboard box (nonliving entity: slow down and nudge) and a flat piece of cardboard lying flush on the road (false positive: keep speed or slow down, keep lane) share the same material but differ in whether they obstruct the vehicle. An open manhole (road hazard: slow down and nudge) and steam rising from a manhole vent (false positive: slow down or creep, keep lane) occupy the same road location but differ in physical obstruction. These contrasts are constructed on the same base scene to control for road geometry, lighting, and context, so that the only difference is the affordance of the inserted object. Figure~\ref{fig:diagnostic-contrasts} illustrates three representative pairs. This design directly tests whether a model grounds its action prediction in the object's physical affordance rather than in superficial visual pattern matching.

\begin{figure*}[t]
    \centering
    \includegraphics[width=0.95\linewidth]{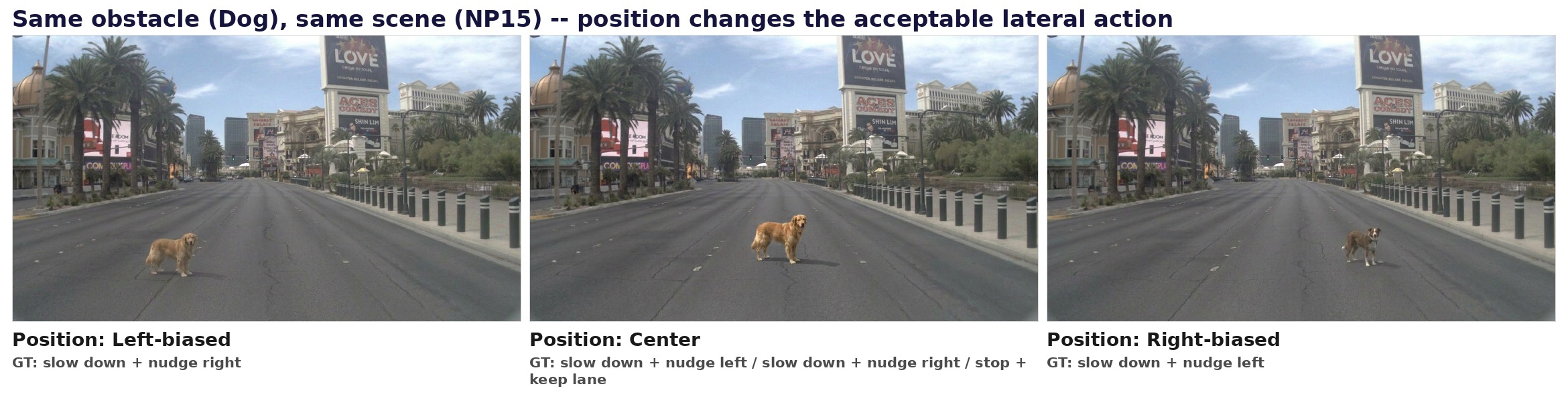}
    \caption{Position-sensitive action variation. The same obstacle (dog) is inserted at three lateral positions in the same base scene. The acceptable lateral action changes with position: left-biased requires nudge right, center accepts either direction, and right-biased requires nudge left.}
    \label{fig:position-sensitive}
\end{figure*}

\paragraph{Position-sensitive action variation.}
Five representative obstacle types---dog (living entity), traffic cone (nonliving entity), pothole (road hazard), flat cardboard (false positive), and plastic bag (false positive)---are inserted at three lateral positions within the ego lane: left-biased, center, and right-biased. The plastic bag is additionally tested in an airborne variant. The same obstacle type on the same base scene produces different acceptable lateral actions depending on its position. For example, a left-biased dog requires nudging right, a center dog accepts nudging in either direction, and a right-biased dog requires nudging left (Figure~\ref{fig:position-sensitive}). This design tests whether the model performs spatial reasoning about obstacle position relative to the ego lane, rather than applying a fixed action template per object category. Full-blockage obstacles are placed across the full lane width and rule out keeping speed; because objects are inserted 15--35 meters ahead, acceptable immediate responses are dominated by deceleration and stopping.

\paragraph{Image generation.}
Counterfactual images are generated using Gemini-3-Pro-Image-Preview, a text-guided image editing model. Before batch generation, we perform \emph{coordinate calibration} for each base scene: annotators manually mark reference coordinates for the left-biased, center, and right-biased positions within the ego lane. These calibrated coordinates serve as the spatial standard for subsequent obstacle insertion, ensuring consistent position definitions across different base scenes rather than relying on the generation model's interpretation of terms such as ``center of lane'' or ``left-biased.'' For each planned sample, the model receives the base scene image and a coordinate prompt specifying the obstacle type, its calibrated position in the ego lane, and the instruction to keep all other scene elements unchanged. We do not use segmentation masks as the primary generation path. Each generation attempt starts from the original base image; failed or low-quality outputs are regenerated from scratch rather than iteratively edited, to avoid visual drift. The editing prompt is structured to preserve road layout, lane markings, vehicles, buildings, and lighting, while inserting a single realistic obstacle at the specified location.

\begin{figure}[h]
    \centering
    \includegraphics[width=\linewidth]{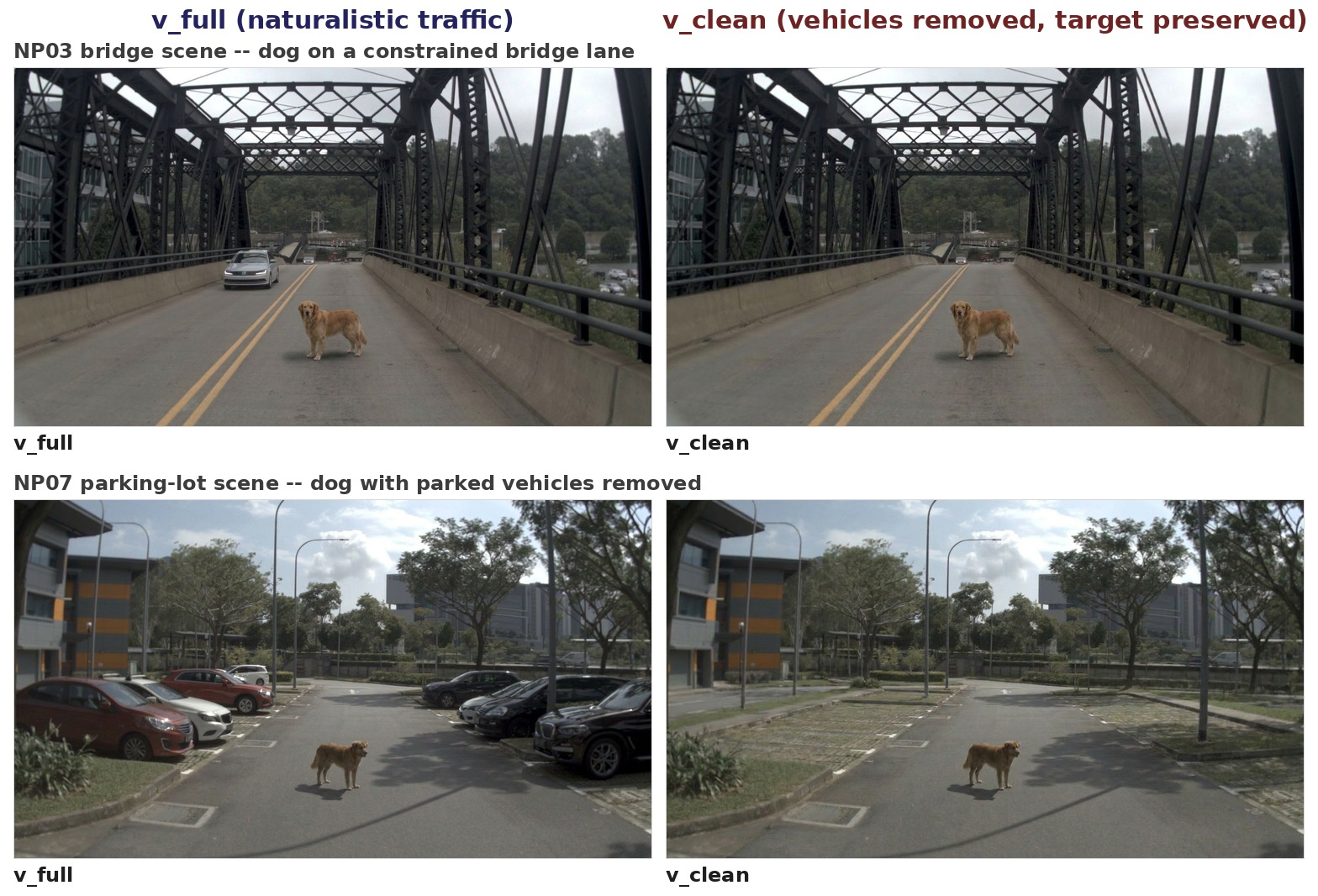}
    \caption{$v_{\mathrm{full}}$ (left) and $v_{\mathrm{clean}}$ (right) versions of the same sample. Background vehicles are removed in $v_{\mathrm{clean}}$ while the target obstacle and road geometry are preserved. This paired design diagnoses whether models rely on traffic-context shortcuts.}
    \label{fig:vfull-vclean-example}
\end{figure}

\paragraph{$v_{\mathrm{full}}$ and $v_{\mathrm{clean}}$ paired construction.}
Each counterfactual sample is produced in two versions. The $v_{\mathrm{full}}$ version preserves surrounding traffic participants (parked cars, moving vehicles) from the original base scene. The $v_{\mathrm{clean}}$ version is derived from the \emph{verified} $v_{\mathrm{full}}$ image by removing background vehicles while preserving the inserted target obstacle, road geometry, and static scene elements. This paired construction serves as a context-cue diagnostic: if a model performs substantially differently between $v_{\mathrm{full}}$ and $v_{\mathrm{clean}}$, it may be relying on surrounding traffic cues (e.g., a lead vehicle already decelerating) rather than directly grounding its decision in the inserted object's affordance. For obstacle types that visually resemble vehicles (e.g., fallen bicycle, fire truck, overturned forklift), the vehicle-removal prompt explicitly identifies the target obstacle and instructs the model not to remove it. Figure~\ref{fig:vfull-vclean-example} shows two examples.

\paragraph{Action-pair labeling and review.}
Each sample is labeled with an acceptable set $\mathcal{P}(x)$ of longitudinal--lateral action pairs rather than a single deterministic label. This design reflects the fact that multiple high-level actions can be safe in a given scene.

The 1,768 $v_{\mathrm{full}}$ images are labeled by three human reviewers. Reviewers follow a shared labeling guide and inspect the edited image, obstacle type, obstacle position, ego-motion history, and navigation command. They assign acceptable action pairs independently, then discuss disagreements to produce the final consensus label set. During adjudication, reviewers remove unsafe lateral directions, add conservative alternatives when appropriate, and relabel spatially displaced obstacles as \emph{keep speed, keep lane} when the inserted object does not occupy the ego vehicle's path.

The final acceptable-pair set for each $v_{\mathrm{full}}$ sample reflects this three-reviewer consensus process; the paired $v_{\mathrm{clean}}$ labels are derived from it as described below.

\paragraph{Annotation reliability.}
\label{app:annotation-reliability}
We define exact agreement strictly: all three annotators must provide identical acceptable sets, and any difference, including one additional acceptable pair, counts as disagreement. On the $v_{\mathrm{full}}$ split, the annotators produce identical sets for 78.51\% of samples, while 21.49\% (380/1,768) require adjudication. Because the labels are set-valued, we report Krippendorff's $\alpha$ with MASI distance, which accounts for partially overlapping and nested sets (Table~\ref{tab:annotation-reliability}). The lower agreement for road hazards reflects genuine ambiguity among cautious lane keeping, lateral avoidance, and stopping. All 380 disagreements are adjudicated pair by pair with reference to obstacle position, available road space, surrounding traffic, ego-motion history, and the navigation command; we do not automatically take the union or intersection. Frequent ambiguities involve response severity, whether lane keeping or lateral avoidance is appropriate, the safe nudge direction, and whether a passable object warrants cautious deceleration.

The paired $v_{\mathrm{clean}}$ labels are constructed differently and therefore do not carry a separate three-annotator agreement statistic. Each $v_{\mathrm{clean}}$ label starts from the adjudicated $v_{\mathrm{full}}$ set and may only add actions that become safe after vehicle removal. Because these additions depend on the road space exposed in the 29 base scenes, the annotation team jointly reviewed the 29 scene geometries and applied the resulting extensions consistently to all derived variants; the subset relation is verified for every pair.

\begin{table}[h]
\centering
\small
\renewcommand{\arraystretch}{1.08}
\begin{tabularx}{\linewidth}{Ycc}
\toprule
Category & Disagreement & $\alpha_{\mathrm{MASI}}$ \\
\midrule
Living entity & 20.69\% & 0.817 \\
Nonliving entity & 18.73\% & 0.815 \\
Road hazard & 35.63\% & 0.692 \\
Full blockage & 16.55\% & 0.783 \\
False positive & 16.32\% & 0.771 \\
\midrule
Overall & 21.49\% & 0.825 \\
\bottomrule
\end{tabularx}
\caption{Annotation reliability on the $v_{\mathrm{full}}$ split: pre-adjudication exact-disagreement rate and set-valued Krippendorff's $\alpha$ with MASI distance, by affordance category.}
\label{tab:annotation-reliability}
\end{table}

\begin{table}[h]
\centering
\small
\renewcommand{\arraystretch}{1.08}
\begin{tabularx}{\linewidth}{YY}
\toprule
Property & Value \\
\midrule
Total reviewed samples & 3,536 \\
Base scenes & 29 (9 Alpamayo + 20 nuPlan mini) \\
Obstacle types & 50 \\
Affordance categories & 5 \\
Versions per sample & $v_{\mathrm{full}}$ and $v_{\mathrm{clean}}$ (1,768 each) \\
\midrule
\multicolumn{2}{l}{\emph{Per-category sample counts ($v_{\mathrm{full}}$)}} \\
\quad Living entity & 348 \\
\quad Nonliving entity & 347 \\
\quad Road hazard & 348 \\
\quad Full blockage & 290 \\
\quad False positive & 435 \\
\midrule
Position variants & center (1,159), left-biased (145), right-biased (145), full-lane blockage (290), airborne (29) \\
Diagnostic contrast pairs & 6 visually matched affordance contrasts \\
Image generation model & Gemini-3-Pro-Image-Preview \\
QA stages & Template policy $\to$ VLM sanity check $\to$ 6 rounds human review \\
\bottomrule
\end{tabularx}
\caption{Summary statistics of \benchmark. Sample counts are for the $v_{\mathrm{full}}$ split; the $v_{\mathrm{clean}}$ split mirrors the same 1,768 samples with background vehicles removed.}
\label{tab:benchmark-stats}
\end{table}

\section{Benchmark Validity Studies}
\label{app:validity}

\subsection{Human Realism and Plausibility Study}
Five independent raters evaluated 125 category-balanced intervention pairs, including both $v_{\mathrm{full}}$ and $v_{\mathrm{clean}}$ versions, yielding 250 synthetic images (50 per affordance category, covering all 29 base scenes). We additionally included 75 naturally occurring corner-case images from a held-out internal dataset (15 per category, no repeated source scenes or clips). Images were presented blindly and rated on 1--5 scales for visual realism and physical/action plausibility (Table~\ref{tab:human-study}).

\begin{table}[h]
\centering
\small
\renewcommand{\arraystretch}{1.08}
\begin{tabularx}{\linewidth}{Ycc}
\toprule
Images & Realism & Plausibility \\
\midrule
Synthetic (\benchmark) & 4.31 [4.24, 4.38] & 4.47 [4.41, 4.53] \\
Real controls & 4.58 [4.50, 4.66] & 4.63 [4.56, 4.70] \\
\bottomrule
\end{tabularx}
\caption{Blind human ratings (1--5) of visual realism and physical/action plausibility. Confidence intervals for synthetic images are obtained by resampling the 29 base scenes while retaining complete intervention pairs.}
\label{tab:human-study}
\end{table}

Overall, 91.2\% (228/250) of synthetic images receive median scores of at least 4 on both dimensions, and ordinal Krippendorff's $\alpha$ is 0.78 for realism and 0.82 for plausibility. Synthetic images score modestly below real images, so we do not claim perceptual equivalence; the results instead support high task-relevant realism and physical plausibility. Because the study is category-balanced, it also tests whether some categories render substantially more cleanly than others: category-level mean realism ranges from 4.18 to 4.40 and plausibility from 4.33 to 4.60, all above 4, with maximum differences of 0.22/0.27 points on the five-point scales. We interpret this as evidence against a large category-wide rendering-quality gap in the accepted images, not as proof that all categories are equally difficult to generate.

\subsection{Real-Image Control Set}
We evaluate the same models on the 75 real corner-case images. The existing human-reviewed labels were independently mapped to the same acceptable action-pair protocol used for \benchmark. Because compatible ego-motion and navigation metadata are unavailable, both the real controls and a 75-image category-matched $v_{\mathrm{full}}$ subset are evaluated with the same image-only prompt, avoiding an input-modality confound. The synthetic and real rankings are strongly correlated: Spearman's $\rho=0.93$, Kendall's $\tau=0.80$, and 70 of 78 pairwise model orderings are preserved. The same-backbone ordering is also unchanged (Table~\ref{tab:real-control}).

\begin{table}[h]
\centering
\small
\renewcommand{\arraystretch}{1.08}
\begin{tabularx}{\linewidth}{Yc}
\toprule
Model & Real-control accuracy \\
\midrule
Qwen3-VL-2B + \method & 61.3\% (46/75) \\
Qwen3-VL-2B + SLERP & 52.0\% (39/75) \\
Qwen3-VL-2B (pretrained) & 49.3\% (37/75) \\
Qwen3-VL-2B + LoRA-SFT & 30.7\% (23/75) \\
\bottomrule
\end{tabularx}
\caption{Same-backbone accuracy on the 75-image real corner-case control set.}
\label{tab:real-control}
\end{table}

Given the size of this control set, we do not claim that every adjacent difference---particularly SLERP versus the pretrained 2B model---is statistically resolved. The robust conclusion is that \method remains the strongest same-backbone variant on previously unseen real scenes. This control set cannot be released because of data-access restrictions; the restriction applies only to this auxiliary evaluation and does not affect the release of the complete public \benchmark.

\subsection{Source-Stratified and Leave-Source-Out Evaluation}
\label{app:source-stratified}
Complete training-data inventories are unavailable for several evaluated models, so individual-image overlap with undisclosed pretraining corpora cannot be ruled out. We therefore measure sensitivity to scene source directly. Each model is evaluated on three subsets: all 3,536 benchmark images; the 2,440 images derived from the 20 nuPlan base scenes (nuPlan only); and the 1,096 images derived from the nine Alpamayo-R1 base scenes (Alpamayo only). Each source-specific subset includes both $v_{\mathrm{full}}$ and $v_{\mathrm{clean}}$ (Table~\ref{tab:source-stratified}).

\begin{table}[h]
\centering
\small
\setlength{\tabcolsep}{4pt}
\renewcommand{\arraystretch}{1.08}
\begin{tabularx}{\linewidth}{Yccc}
\toprule
Model & All & nuPlan & Alpamayo \\
\midrule
GPT-5.5 & 83.0 & 84.4 & 79.8 \\
Qwen3-VL-8B & 65.5 & 65.9 & 64.6 \\
Qwen3-VL-2B + \method & 60.8 & 60.6 & 61.1 \\
Alpamayo-1.5-10B & 59.3 & 60.5 & 56.5 \\
Qwen3-VL-32B & 59.0 & 60.7 & 55.3 \\
Qwen3-VL-4B & 56.9 & 56.8 & 57.0 \\
Qwen3-VL-2B + SLERP & 53.0 & 52.6 & 53.9 \\
Qwen3-VL-2B & 50.3 & 49.0 & 53.3 \\
AutoDrive-R$^2$-7B & 48.6 & 47.5 & 51.2 \\
Cosmos-Reason2-8B & 37.1 & 39.3 & 32.2 \\
Cosmos-Reason2-2B & 34.9 & 34.3 & 36.3 \\
Qwen3-VL-2B + LoRA-SFT & 32.4 & 29.6 & 38.6 \\
\bottomrule
\end{tabularx}
\caption{Source-stratified pair accuracy: all scenes, the nuPlan-only subset (2,440 images), and the Alpamayo-only subset (1,096 images). Curious-VLA-3B is omitted because its outputs are largely incompatible with the structured decision interface (Appendix~\ref{app:prompting-parsing}).}
\label{tab:source-stratified}
\end{table}

Three findings address the fairness concern. First, Alpamayo-1.5-10B shows no home-source advantage: it scores 56.5 on the Alpamayo-only subset and 60.5 on the nuPlan-only subset, the opposite of the pattern expected if familiarity with the released Alpamayo scenes materially inflated its score. Familiarity with an original base scene would also not directly reveal the benchmark label, because every sample contains a counterfactually inserted object whose identity and placement determine the acceptable action set. Second, the main \method conclusions are unchanged after removing all Alpamayo-sourced scenes: on the nuPlan-only subset, \method retains gains of 11.6, 31.0, and 8.0 points over the pretrained backbone, LoRA-SFT, and SLERP, respectively. Third, \method is nearly invariant to scene source (60.6 versus 61.1). Because \method, Alpamayo-1.5-10B, and Qwen3-VL-32B are separated by at most 0.2 points on the nuPlan-only subset, we avoid drawing conclusions from their relative ordering there.

\section{In-Domain Driving Split}
\label{app:indomain-split}

\paragraph{In-domain driving data and splits.}
The in-domain corpus contains 10,000 multi-camera driving clips spanning 26 ODD categories. A 90/10 clip-level, multi-label-stratified split produces 9,002 training clips (199,802 samples) and 998 held-out test clips (21,857 samples). The LoRA driving expert and \regmoe adapters are trained only on the 199,802 training samples. Nominal driving performance is evaluated on a 3,613-sample subset drawn exclusively from the held-out test clips. This subset covers all 998 test clips, all 26 ODD categories, and all 72 action labels, while deviating from the full held-out distribution by at most 0.24 percentage points per action label and 0.68 percentage points per ODD category. Because splitting is performed at the clip level, no clip or frame overlaps between training and nominal evaluation. \benchmark is used exclusively for evaluation and is disjoint from the in-domain training data. Each evaluation sample consists of a front-view driving image, recent ego-motion history, a navigation command, and one canonical longitudinal--lateral action pair. Table~\ref{tab:indomain-summary} summarizes the split.

\begin{table}[h]
\centering
\small
\renewcommand{\arraystretch}{1.08}
\begin{tabularx}{\linewidth}{YY}
\toprule
Property & Value \\
\midrule
Training clips / samples & 9,002 / 199,802 \\
Held-out test clips / samples & 998 / 21,857 \\
Nominal evaluation subset & 3,613 samples (from held-out clips only) \\
Input modality & Front-view image, ego-motion history, navigation command \\
Output label & One canonical longitudinal--lateral action pair \\
Usage of the 3,613 subset & Held-out nominal evaluation only \\
Relation to \benchmark & Nominal driving split, not counterfactual intervention benchmark \\
\bottomrule
\end{tabularx}
\caption{Summary of the in-domain driving data and splits.}
\label{tab:indomain-summary}
\end{table}

\paragraph{Action distribution.}
Because nominal driving data is often dominated by frequent behaviors such as keeping lane and maintaining speed, we report the longitudinal and lateral action distributions in Table~\ref{tab:indomain-action-dist}. This distribution is useful for interpreting the gap between in-domain performance and \benchmark performance: a model may achieve high nominal accuracy by learning frequent driving priors, while still failing under rare-object interventions.

\begin{table}[h]
\centering
\small
\renewcommand{\arraystretch}{1.08}
\begin{tabularx}{\linewidth}{Ycc}
\toprule
Action type & Action & Count / Percentage \\
\midrule
Longitudinal & keep speed & 1{,}440 / 39.9\% \\
Longitudinal & slow down & 1{,}172 / 32.4\% \\
Longitudinal & yield & 428 / 11.8\% \\
Longitudinal & creep & 81 / 2.2\% \\
Longitudinal & stop & 492 / 13.6\% \\
\midrule
Lateral & keep lane & 2{,}564 / 71.0\% \\
Lateral & nudge left & 388 / 10.7\% \\
Lateral & nudge right & 115 / 3.2\% \\
Lateral & lane change left & 428 / 11.8\% \\
Lateral & lane change right & 118 / 3.3\% \\
\bottomrule
\end{tabularx}
\caption{Action distribution of the in-domain driving split. The distribution helps clarify whether nominal evaluation is dominated by frequent driving behaviors.}
\label{tab:indomain-action-dist}
\end{table}

\paragraph{Separation from \benchmark.}
$\mathcal{D}_{\mathrm{in}}$ and \benchmark serve different roles. $\mathcal{D}_{\mathrm{in}}$ provides nominal driving supervision and evaluation, while \benchmark evaluates controlled rare-object interventions. We ensure that \benchmark is not treated as additional supervised training data for the main adaptation pipeline. This separation allows us to test whether a model trained on nominal driving data can generalize to counterfactual long-tail affordance prediction.

\paragraph{Interpretation.}
The in-domain split should not be interpreted as a long-tail benchmark. Its role is to establish whether a model can perform ordinary driving meta-action prediction after adaptation. The contrast between $\mathcal{D}_{\mathrm{in}}$ and \benchmark is central to our evaluation: high in-domain accuracy with low \benchmark accuracy indicates over-specialization to frequent driving priors, whereas improvement on both splits suggests better driving-specific action grounding.

\section{Action Space and Labeling Protocol}
\label{app:labeling}

\paragraph{Action space.}
The model outputs one longitudinal action and one lateral action. The longitudinal action captures high-level speed-control behavior, while the lateral action captures high-level lane-position behavior. All labels follow the same immediate-response semantics: the acceptable set $\mathcal{P}(x)$ describes the next high-level action under the current observation, not the vehicle's final state. Table~\ref{tab:action-space} summarizes the action space used in our benchmark.

\begin{table}[h]
\centering
\small
\renewcommand{\arraystretch}{1.08}
\begin{tabularx}{\linewidth}{YY}
\toprule
Action type & Candidate actions \\
\midrule
Longitudinal &
keep speed; slow down; yield; creep; stop \\
Lateral &
keep lane; nudge left; nudge right; lane change left; lane change right \\
\bottomrule
\end{tabularx}
\caption{Longitudinal and lateral meta-action space. The final prediction is evaluated as a pair rather than as two independent labels.}
\label{tab:action-space}
\end{table}

\paragraph{Acceptable action-pair sets.}
For each sample $x$, the reference label is an acceptable set $\mathcal{P}(x)$ rather than a single action pair. This design reflects the fact that multiple high-level actions can be safe or semantically equivalent in a given scene. For example, a living entity in the ego lane may accept both \emph{yield, keep lane} and \emph{stop, keep lane}, while a left-biased rigid obstacle may accept \emph{slow down, nudge right}. Directionally wrong or unsafe maneuvers are excluded from $\mathcal{P}(x)$.

Table~\ref{tab:set-size} reports how acceptable-set size varies across categories. Full blockage has the smallest acceptable sets yet the relatively high \method accuracy (69.7\%), and false positives share the largest mean set size with living entities but behave differently relative to SLERP; acceptable-set size therefore does not explain the category performance pattern.

\begin{table}[h]
\centering
\small
\renewcommand{\arraystretch}{1.08}
\begin{tabularx}{\linewidth}{Ycc}
\toprule
Category & Mean set size & Median \\
\midrule
Living entity & 1.67 & 1 \\
Nonliving entity & 1.44 & 1 \\
Road hazard & 1.45 & 1 \\
Full blockage & 1.31 & 1 \\
False positive & 1.67 & 2 \\
\bottomrule
\end{tabularx}
\caption{Acceptable action-pair set size by affordance category.}
\label{tab:set-size}
\end{table}

\paragraph{Label generation and review.}
We label each sample in three stages. First, an obstacle- and position-conditioned template policy assigns candidate longitudinal--lateral action pairs. Second, a VLM-based sanity check identifies inconsistent or ambiguous labels, such as cases where the proposed lateral direction conflicts with the object position. Third, human review verifies the final acceptable action-pair set. This multi-stage process is designed to preserve reasonable action ambiguity while filtering out unsafe or directionally invalid decisions.

\section{Prompting and Parsing Protocol}
\label{app:prompting-parsing}

\paragraph{Structured prompt.}
All models are evaluated using the same structured perception-to-decision prompt. The prompt asks the model to identify decision-relevant objects, infer spatial, physical, and normative constraints, and output a final longitudinal--lateral action pair in a canonical format. This design reduces variation caused by free-form explanation style and focuses evaluation on the final decision interface.

\paragraph{Prompt template.}
The following template is used for evaluation:

\begin{quote}
\small
Given the front-view driving image, recent ego-motion history, and navigation command, identify the decision-relevant object or hazard. Determine whether it changes the ego vehicle's feasible high-level action space. Then output the final decision using one longitudinal action and one lateral action.

Longitudinal action must be one of: keep speed, slow down, yield, creep, stop.

Lateral action must be one of: keep lane, nudge left, nudge right, lane change left, lane change right.

Final answer format: longitudinal action = [action]; lateral action = [action].
\end{quote}

\paragraph{Decoding configuration.}
All models are evaluated with greedy decoding, repetition penalty $1.1$, and a maximum generation length of 1{,}024 tokens.

\paragraph{Extraction, normalization, and scoring.}
Evaluation uses a three-stage pipeline applied identically to all models. (1)~\emph{Rule-based extraction}: a deterministic parser isolates the model's committed decision span from the raw response; if the response contains multiple candidate actions, the explicitly marked final answer is used, and if no unique decision span can be recovered, the output is marked invalid. (2)~\emph{LLM decision normalization}: a text-only LLM normalizer (DeepSeek-v4-Pro, temperature~$0$) reads only the extracted text---not the image or reference labels---and maps it to the canonical longitudinal--lateral action pair, so that semantically equivalent non-canonical phrasings are normalized consistently across models. (3)~\emph{Rule-based scoring}: the normalized pair is scored deterministically against the acceptable set $\mathcal{P}(x)$; the LLM does not determine correctness. Invalid outputs are counted as incorrect. This evaluation component is distinct from the VLM-based sanity check used during dataset construction (Appendix~\ref{app:labeling}).

\paragraph{Invalid-output rates.}
Table~\ref{tab:invalid-rates} reports invalid-output rates on the $v_{\mathrm{full}}$ split. Curious-VLA-3B requires separate interpretation because it is post-trained as a trajectory-generating VLA: 1,607 of its 1,690 invalid outputs contain multiple candidate actions without committing to one final pair, while only 83 contain no recoverable action pair. Its score therefore measures end-to-end compatibility with the required decision interface rather than its underlying driving reasoning.

\begin{table}[h]
\centering
\small
\renewcommand{\arraystretch}{1.08}
\begin{tabularx}{\linewidth}{Ycc}
\toprule
Model & Invalid rate & Pair Acc. \\
\midrule
GPT-5.5 & 0.68\% & 82.2 \\
Alpamayo-1.5-10B & 0.17\% & 59.8 \\
Qwen3-VL-8B & 0.28\% & 65.1 \\
Qwen3-VL-2B + \method & 4.30\% & 60.1 \\
AutoDrive-R$^2$-7B & 14.76\% & 48.1 \\
Cosmos-Reason2-2B & 19.91\% & 35.6 \\
Cosmos-Reason2-8B & 28.28\% & 41.0 \\
Curious-VLA-3B & 95.59\% & 1.2 \\
\bottomrule
\end{tabularx}
\caption{Invalid-output rates and pair accuracy on the $v_{\mathrm{full}}$ split under the shared three-stage scoring pipeline.}
\label{tab:invalid-rates}
\end{table}

\section{Language-Side Knowledge Retention Analysis}
\label{app:retention}

Long-tail affordance prediction relies on two types of knowledge: visual object representations and language-side situational reasoning that maps objects to driving implications. Since all adaptation variants freeze the ViT encoder, a linear-probe check on the visual encoder gives identical open-world visual accuracy across checkpoints. We therefore treat this result only as a sanity check that the visual backbone is not modified, rather than as a discriminative measure of knowledge retention.

To evaluate whether adaptation changes the model's reasoning ability, we further use a text-based retention probe constructed from DriveQA-T~\citep{wei2025driveqa}, a US driving-exam benchmark. Specifically, we curate 1{,}292 situational driving-reasoning questions and evaluate each model with a 4-choice multiple-choice protocol. To reduce answer-position bias, we report rotation-averaged accuracy over answer-choice permutations. This probe tests whether driving adaptation preserves the pretrained model's situational reasoning needed for rare-object and safety-critical scenarios.

Table~\ref{tab:retention} shows that direct LoRA SFT substantially hurts retention, dropping from 79.6\% to 74.1\%. SLERP partially mitigates this degradation but still remains 4.4 percentage points below the pretrained model. In contrast, \method{} achieves 79.1\% retention, only 0.5 percentage points below the pretrained backbone, while also improving \benchmark accuracy from 50.3\% to 60.8\%. These results support the intended design of \method: adapting the model toward driving-specific affordance prediction without substantially erasing its pretrained situational reasoning ability. These experiments do not establish retention across general VQA or broad multimodal perception tasks; we therefore restrict our claims to language-side knowledge retention, and general multimodal retention remains outside the scope of the current evaluation.

\begin{table}[t]
\centering
\small
\setlength{\tabcolsep}{4pt}
\renewcommand{\arraystretch}{1.10}
\begin{tabularx}{\linewidth}{Yccc}
\toprule
Model & \shortstack{\benchmark\\Acc.} & \shortstack{DriveQA-T\\Retention} & $\Delta$ Ret. \\
\midrule
Pretrained 2B VLM     & 50.3 & 79.6 & 0.0 \\
LoRA SFT              & 32.4 & 74.1 & -5.5 \\
SLERP merge           & 53.0 & 75.2 & -4.4 \\
\method{}             & 60.8 & 79.1 & -0.5 \\
\bottomrule
\end{tabularx}
\caption{Knowledge retention analysis. Retention is the rotation-averaged 4-choice multiple-choice accuracy (\%) on 1{,}292 situational driving-reasoning questions curated from DriveQA-T~\citep{wei2025driveqa}; chance performance is 25.0\%. All variants freeze the ViT encoder, and a separate linear-probe sanity check gives identical visual-feature accuracy across checkpoints.}
\label{tab:retention}
\end{table}

\section{Exploratory Adaptation Variants}
\label{app:exploratory-adaptation}

Before arriving at the final \regmoe design, we explored two additional adaptation directions: Fisher-guided gated adapters and action-grounded GRPO. These variants were motivated by the same semantic-action tension as \method: the model should acquire driving-specific meta-action behavior without overwriting the open-world knowledge needed for rare-object reasoning. We report them as exploratory experiments to clarify why the final design emphasizes regime-aware capacity allocation rather than sensitivity-based adapter placement or direct action-level reward optimization.

\subsection{Fisher-Guided Gated Expert Adapters}

We first explored whether adapter placement could be improved by avoiding layers that are important for general vision-language knowledge. Starting from the SLERP-merged backbone $\theta^{\star}$, we estimated layer sensitivity using a diagonal empirical Fisher score on a generic vision-language calibration set:
\[
    \operatorname{Imp}(\ell)
    =
    \sum_{i\in\ell}
    \mathbb{E}
    \left[
    \left(
    \nabla_{\theta_i}
    \log p_{\theta^{\star}}(z\mid I,x)
    \right)^2
    \right].
\]
Layers with high Fisher scores were treated as knowledge-sensitive, and adapters were placed in lower-sensitivity layers under a spacing constraint. For each selected layer $\ell$, we added a gated residual adapter:
\[
    h'_{\ell}
    =
    h_{\ell}
    +
    g_{\ell}(h_{\ell}) \odot E_{\ell}(h_{\ell}).
\]
Although this design provides a conservative way to inject adaptation capacity, it does not improve over the SLERP baseline on \benchmark. This indicates that avoiding knowledge-sensitive layers is insufficient for counterfactual affordance prediction: the key challenge is not only preserving generic representations, but assigning different adaptation directions to different driving decision regimes. This observation motivated the regime-aware routing design in \regmoe.

\subsection{Action-Grounded GRPO}

We also explored action-grounded GRPO to directly optimize parsed action correctness. For each input, the policy samples a group of responses. Each response is parsed into a longitudinal--lateral action pair, and the reward is computed from format validity and action-pair correctness:
\[
R_i
=
\mathbb{I}_{\mathrm{fmt}}(o_i)
\left(
w_{\mathrm{lon}}m_{\mathrm{lon}}
+
w_{\mathrm{lat}}m_{\mathrm{lat}}
\right)
-
\mathbb{I}_{\neg \mathrm{fmt}}(o_i),
\]
where $m_{\mathrm{lon}}$ and $m_{\mathrm{lat}}$ indicate whether the parsed longitudinal and lateral actions are acceptable, and $\mathbb{I}_{\mathrm{fmt}}(o_i)$ indicates whether the output follows the required action format. The group-relative advantage is
\[
    A_i =
    \frac{
    R_i-\operatorname{mean}(\{R_j\}_{j=1}^{G})
    }{
    \operatorname{std}(\{R_j\}_{j=1}^{G})+\epsilon
    }.
\]
Although this objective is better aligned with the parsed evaluation metric than token likelihood, it introduces additional optimization sensitivity and does not consistently improve counterfactual affordance accuracy. In our experiments, GRPO slightly improves in-domain accuracy but leaves \benchmark accuracy nearly unchanged, suggesting that direct reward optimization alone cannot resolve the underlying regime-mixing problem. We therefore keep GRPO as an exploratory variant rather than part of the main pipeline.

\begin{table}[h]
\centering
\small
\renewcommand{\arraystretch}{1.08}
\begin{tabularx}{\linewidth}{Ycc}
\toprule
Variant & In-domain Acc. & \benchmark Acc. \\
\midrule
SLERP merge & 55.9 & 53.0 \\
SLERP + Fisher adapter & 54.0 & 52.8 \\
SLERP + Fisher adapter + GRPO & 56.5 & 52.9 \\
SLERP + \regmoe & 52.8 & 60.8 \\
\bottomrule
\end{tabularx}
\caption{Exploratory adaptation variants. Fisher-guided adapters and GRPO were evaluated during method development, but \regmoe provides the strongest counterfactual affordance accuracy.}
\label{tab:exploratory-adaptation}
\end{table}

Table~\ref{tab:exploratory-adaptation} shows that neither exploratory direction provides the desired long-tail gain. Fisher-guided adapters slightly reduce \benchmark accuracy relative to the SLERP merge, suggesting that sensitivity-guided placement alone does not provide the right specialization structure. Adding GRPO improves in-domain accuracy but leaves \benchmark accuracy nearly unchanged. In contrast, \regmoe trades a small amount of in-domain accuracy for substantially higher \benchmark accuracy, supporting our hypothesis that counterfactual affordance prediction benefits more from regime-aware expert routing than from conservative adapter placement or direct action-level reward optimization.

\section{Implementation Details}
\label{app:implementation}

\paragraph{Backbone and adaptation.}
Our main backbone is Qwen3-VL-2B. We first train a LoRA driving expert on the in-domain driving split, then materialize the LoRA update into the base model to obtain a driving expert. The expert is merged with the pretrained model using SLERP. We then freeze the merged backbone and train \regmoe adapters. The final checkpoint stores only the MoE-specific parameters, including LoRA expert matrices, router weights, and regime-conditioned routing biases. Fisher-guided adapters and GRPO-refined variants are described separately in Appendix~\ref{app:exploratory-adaptation}.

\paragraph{RegMoE configuration.}
For Qwen3-VL-2B, \regmoe is inserted into layers $\{18{:}27\}$ and adapts the $q$, $k$, $v$, $o$, gate, up, and down projections, resulting in 70 adapted projection sites. We use 6 experts, LoRA rank $r=8$, and LoRA scaling $\alpha/r=16/8=2$.

\paragraph{Routing regularizers.}
The load-balancing term is
\[
    \mathcal{L}_{\mathrm{lb}}
    =
    \left\|
    \bar{\boldsymbol{\pi}}
    -
    \frac{1}{E}\mathbf{1}
    \right\|_2^2,
    \qquad
    \bar{\boldsymbol{\pi}}
    =
    \frac{1}{N}
    \sum_{i=1}^{N}
    \boldsymbol{\pi}(x_i,t_i).
\]
The task-separation term is
\[
    \mathcal{L}_{\mathrm{sep}}
    =
    \frac{1}{E}
    \sum_{e=1}^{E}
    \operatorname{Var}_{t}
    \left(
    \bar{\pi}_{t,e}
    \right),
\]
where $\bar{\pi}_{t,e}$ is the average routing probability of expert $e$ under task type $t$.

\paragraph{Training hyperparameters.}
Table~\ref{tab:training-hyperparams} lists the main training hyperparameters. We keep the table in the appendix because these details are important for reproducibility but not central to the main story.

\begin{table}[h]
\centering
\small
\renewcommand{\arraystretch}{1.08}
\begin{tabularx}{\linewidth}{YY}
\toprule
Hyperparameter & Value \\
\midrule
Backbone & Qwen3-VL-2B \\
SLERP coefficient $\alpha$ & 0.08 \\
Learning rate & $5 \times 10^{-5}$ \\
Batch size & 128 \\
Training epochs & 1 \\
Number of \regmoe experts & 6 \\
\regmoe layers & 10 late layers \\
Adapted projection modules & 7 \\
LoRA rank & 8 \\
LoRA alpha & 16 \\
Answer-token weight $\lambda_{\mathrm{ans}}$ & 5.0 \\
Load-balancing weight $\lambda_{\mathrm{lb}}$ & 0.1 \\
Separation weight $\lambda_{\mathrm{sep}}$ & 0.05 \\
\bottomrule
\end{tabularx}
\caption{Training hyperparameters used for the final \method model.}
\label{tab:training-hyperparams}
\end{table}

\section{Seed Variance and Parameter-Matched Comparison}
\label{app:seed-variance}

To test whether \regmoe's improvement can be explained by additional parameter capacity or run variance, we compare it against a parameter-matched single-adapter baseline (E1a): one rank-48 LoRA over the same seven projection types and ten layers, matching the total rank of six rank-8 experts. This baseline is distinct from the ``Single LoRA adapter'' ablation in Table~\ref{tab:ablation}, which is not parameter-matched to \regmoe. Both E1a and \regmoe use the same frozen SLERP backbone, training data, and optimization settings. Table~\ref{tab:seed-variance} reports $v_{\mathrm{full}}$ accuracy across four pre-specified seeds.

\begin{table}[h]
\centering
\small
\renewcommand{\arraystretch}{1.08}
\begin{tabularx}{\linewidth}{Yccc}
\toprule
Seed & \regmoe & E1a & Paired $\Delta$ \\
\midrule
42 & 60.07 & 56.67 & +3.40 \\
10 & 60.52 & 55.77 & +4.75 \\
20 & 58.48 & 54.30 & +4.18 \\
30 & 59.79 & 55.32 & +4.47 \\
\midrule
Mean $\pm$ SD & 59.72 $\pm$ 0.88 & 55.52 $\pm$ 0.99 & +4.20 $\pm$ 0.58 \\
\bottomrule
\end{tabularx}
\caption{$v_{\mathrm{full}}$ accuracy of \regmoe and the parameter-matched single-adapter baseline (E1a) across four seeds.}
\label{tab:seed-variance}
\end{table}

\regmoe outperforms E1a under every seed; the mean paired gap is 4.20 points with a t-based 95\% confidence interval of $[3.27, 5.13]$, and seed-to-seed variation ($\approx$1 point per method) is substantially smaller than the paired gap. The regime label is not an additional action-prediction target or auxiliary output loss; it supervises only the router during training, so a single adapter has no equivalent router to which the same supervision could be applied. This comparison therefore supports the benefit of the complete \regmoe mechanism---training-time action-label-guided routing together with multiple low-rank experts---relative to an equally sized non-routed adapter, and we do not attribute the improvement to expert multiplicity independently of its routing supervision.

\section{Efficiency Profiling}
\label{app:efficiency}

We profile inference overhead on an NVIDIA H100 with batch size 1 under the same precision, input, and decoding configuration used in our experiments (Table~\ref{tab:efficiency}).

\begin{table}[h]
\centering
\small
\setlength{\tabcolsep}{3pt}
\renewcommand{\arraystretch}{1.08}
\begin{tabularx}{\linewidth}{Yccc}
\toprule
Model & \shortstack{Added\\params} & \shortstack{Peak\\memory} & \shortstack{Decoding\\throughput} \\
\midrule
SLERP-merged 2B backbone & 0 & $\approx$4.3 GB & 33.4 tok/s \\
Single rank-48 LoRA (unmerged) & 19.8M & $\approx$4.4 GB & 21.0 tok/s \\
\regmoe (fused) & 19.8M & $\approx$4.5 GB & 22.4 tok/s \\
\bottomrule
\end{tabularx}
\caption{Inference profiling on an NVIDIA H100 (batch size 1) under the paper's inference configuration.}
\label{tab:efficiency}
\end{table}

\regmoe adds 19.8M parameters (0.93\% of the 2.13B-parameter backbone), with a peak-memory overhead of approximately 0.23~GB over the merged backbone and 0.10~GB over the parameter-matched single-adapter implementation. The fused implementation stacks the six expert projections into two matrix multiplications before applying the input-dependent routing weights; this is functionally equivalent to evaluating the experts separately but avoids repeated kernel launches, reaching 22.4 tokens/s (44.6 ms per decoded token) versus 33.4 tokens/s (29.9 ms/token) for the adapter-free backbone. A fixed single LoRA can be permanently merged into its backbone and can therefore approach backbone throughput in deployment; \regmoe cannot be permanently merged because its expert mixture is input-dependent. We accordingly claim input-dependent specialization with less than 1\% parameter overhead and approximately 0.2~GB additional peak memory, not a latency advantage over a merged single adapter.

Under the same profiling setup, \method improves \benchmark accuracy from 50.3\% to 60.8\% with an approximately 4.5~GB footprint, whereas Qwen3-VL-8B reaches 65.5\% but requires 16.8~GB: \method recovers approximately 69\% of the accuracy gap between the zero-shot 2B and 8B models while remaining within the memory class of the 2B backbone. Because these measurements are obtained on an H100 rather than an automotive accelerator, they quantify relative overhead but do not establish direct in-vehicle deployability.

\section{Error Analysis}
\label{app:error-analysis}

\subsection{Predicted Action Distributions}
\label{app:action-distribution}
Table~\ref{tab:action-distribution} compares \method's predicted longitudinal action distributions on nominal driving and \benchmark. On nominal driving, \method keeps speed on nearly half of the samples and its stop rate closely matches the ground-truth rate (13.8\% versus 13.6\%); a constant \emph{slow down, keep lane} policy achieves only 24.7\% nominal accuracy, compared with \method's 52.8\%. On \benchmark, predictions shift primarily from \emph{keep speed} to \emph{slow down} rather than toward \emph{stop} or \emph{yield}, consistent with cautious initial deceleration when inserted objects occupy or approach the ego path. The contrast between the two distributions indicates an intervention-conditioned response rather than a fixed global conservative bias. Note that meta-action labels denote the immediate response rather than the terminal maneuver: because obstacles are inserted 15--35 meters ahead of the ego vehicle, cautious deceleration (\emph{slow down}) on approach is typically within the acceptable set even for full blockages, with \emph{stop} required only when the scene leaves no approach margin. The low \emph{stop} rate is therefore consistent with the high full-blockage accuracy in Table~\ref{tab:category-results}.

\begin{table}[h]
\centering
\small
\renewcommand{\arraystretch}{1.08}
\begin{tabularx}{\linewidth}{Ycc}
\toprule
Longitudinal prediction & Nominal & \benchmark \\
\midrule
keep speed & 48.3\% & 9.3\% \\
slow down & 28.3\% & 84.7\% \\
yield & 8.5\% & 0.7\% \\
creep & 1.1\% & 4.6\% \\
stop & 13.8\% & 0.7\% \\
\bottomrule
\end{tabularx}
\caption{\method's predicted longitudinal action distribution on nominal driving versus \benchmark.}
\label{tab:action-distribution}
\end{table}

\subsection{Error Taxonomy}
We group representative errors into four categories (Table~\ref{tab:error-taxonomy}). This analysis helps distinguish ordinary perception failures from the semantic-action grounding failures targeted by \benchmark.

\begin{table}[!h]
\centering
\small
\renewcommand{\arraystretch}{1.08}
\begin{tabularx}{\linewidth}{YY}
\toprule
Error type & Description \\
\midrule
Object recognition error &
The model fails to identify the inserted rare object or confuses it with another object. \\
Affordance mapping error &
The model recognizes the object but assigns the wrong driving implication, such as treating a low-risk object as a full blockage. \\
Direction error &
The model predicts a plausible longitudinal action but an unsafe or inconsistent lateral action, such as nudging toward the obstacle. \\
Invalid-format error &
The model fails to produce a parsable longitudinal--lateral action pair. \\
\bottomrule
\end{tabularx}
\caption{Error taxonomy for \benchmark. The taxonomy separates perception failures from affordance mapping and action-format failures.}
\label{tab:error-taxonomy}
\end{table}

\paragraph{Interpretation.}
Object recognition errors indicate that the model lacks sufficient visual grounding for rare objects. Affordance mapping errors are more central to our benchmark: they occur when the object is recognized but its implication for the ego vehicle's feasible action space is wrong. Direction errors reveal failures in spatial grounding, especially when the obstacle position should determine whether the model nudges left or right. Invalid-format errors reflect interface failures and are directly targeted by the structured perception-to-decision prompt.

Table~\ref{tab:error-counts} reports error-type counts over $v_{\mathrm{full}}$ failures. The recognition and affordance columns use a keyword heuristic over full model responses; direction and invalid-format counts are exact. \method's failure profile differs qualitatively from the base model: affordance-mapping errors drop from 50.3\% to 23.0\% of failures, while direction errors rise to 59.1\%, consistent with a model that correctly identifies the need to act but misjudges the lateral direction.

\begin{table}[h]
\centering
\footnotesize
\setlength{\tabcolsep}{2.5pt}
\renewcommand{\arraystretch}{1.08}
\begin{tabularx}{\linewidth}{@{}>{\raggedright\arraybackslash}X rrrrr@{}}
\toprule
Model & Fail. & Recog. & Afford. & Direct. & Invalid \\
\midrule
Base & 907 & 51 & 456 & 375 & 25 \\
+ LoRA-SFT & 1,156 & 900 & 77 & 178 & 1 \\
+ SLERP & 874 & 55 & 415 & 378 & 26 \\
+ \method & 706 & 51 & 162 & 417 & 76 \\
\bottomrule
\end{tabularx}
\caption{Error-type counts over $v_{\mathrm{full}}$ failures for Qwen3-VL-2B variants: object recognition, affordance mapping, direction, and invalid-format errors.}
\label{tab:error-counts}
\end{table}

\end{document}